\documentclass[lettersize,journal]{IEEEtran}
\usepackage{amsmath,amssymb,amsfonts}
\usepackage{algorithmic}
\usepackage{graphicx}
\usepackage{textcomp}
\usepackage{booktabs}
\usepackage{cite} 
\usepackage{amsmath} 
\usepackage{algorithmic} 
\usepackage{array} 
\usepackage{mdwmath} 
\usepackage{mdwtab} 
\usepackage{eqparbox} 
\usepackage[caption=false,font=footnotesize]{subfig} 
\usepackage{dblfloatfix} 
\usepackage{url} 
\usepackage{verbatim} 
\usepackage{balance}
\usepackage{subfig} 

    \usepackage{threeparttable}
    \usepackage{threeparttablex}
    \usepackage{verbatim}
\usepackage[usestackEOL]{stackengine}
\usepackage{xcolor, soul}
\definecolor{codegreen}{rgb}{0,0.6,0}
\definecolor{codegray}{rgb}{0.5,0.5,0.5}
\definecolor{codepurple}{rgb}{0.58,0,0.82}
\definecolor{backcolour}{rgb}{0.95,0.95,0.92}

\definecolor{transparent}{rgb}{1,1,1} 
\definecolor{yellow}{rgb}{1,1,0} 
\DeclareRobustCommand{\hlg}[1]{{\sethlcolor{transparent}\hl{#1}}} %

\begin{document}

\title{MC-QDSNN: Quantized Deep evolutionary SNN with Multi-Dendritic Compartment Neurons for Stress Detection using Physiological Signals}

\author{
\IEEEauthorblockN{
Ajay B S\IEEEauthorrefmark{2}, 
Phani Pavan K\IEEEauthorrefmark{2}, and 
Madhav Rao\IEEEauthorrefmark{2}
\IEEEmembership{Senior Member, IEEE}}
\IEEEauthorblockA{
\IEEEauthorrefmark{2}{International Institute of Information Technology Bangalore, India, (e-mail: 
\{ajay.b, phanipavan.k, mr\}@iiitb.ac.in)}}
\thanks{The work was supported by IIIT-Bangalore, India.}
\thanks{Manuscript received April 11, 2024; revised August 10, 2024.}}




\maketitle

\begin{abstract}
Long short-term memory (LSTM) has emerged as a definitive network model 
for analyzing and inferring time series data since their introduction.
LSTM has the capability to not only extract spectral features similar to convolutional-neural-network~(CNN) models, but also a mixture of temporal features.
Due to this distinguished advantage, similar feature extraction method is explored for the spiking counterpart of the neural network, targeted for time-series data.
Though LSTMs perform well in the spiking form of neural network, they tend to be compute and power intensive. Addressing this issue, the work proposes Multi-Compartment Leaky ({\it MCLeaky}) neuron as a viable alternative for efficient processing of time series data. The {\it MCLeaky} neuron, introduced as a derivative of the Leaky Integrate and Fire~(LIF) neuron model, contains multiple memristive synapses interlinked to form the memory component of the neuron, by emulating Hippocampus’ structure of brain as reference. The
proposed {\it MCLeaky} neuron based Spiking Neural Network model and its quantized variant were benchmarked against
state-of-the-art~(SOTA) Spiking LSTMs 
to perform human stress detection by comparing computing requirements, compute-latency and real-world performances on freshly acquired unseen data with models that is acquired by employing Neural Architecture Search~(NAS). Results show that the networks with {\it MCLeaky} activation neuron managed a superior 
accuracy of 98.8\% 
to detect stress based on Electrodermal activity (EDA) signals, better than any other investigated model, while using 20\% less parameters on average. 
{\it MCLeaky} neuron was also investigated for different modality of signals including EDA Wrist, EDA Chest, Temperature, ECG signal, and combination of them.
Quantized {\it MCLeaky} model 
was also derived and validated to 
forecast their performance on hardware aware architecture, which resulted in 91.84\% accuracy. 
The neurons were evaluated for multiple modalities of data 
towards stress detection, 
which resulted in energy savings of 25.12$\times$ to 39.20$\times$ and EDP gains of
52.37$\times$ to 81.9$\times$ over ANN model, besides offering
the best accuracy of 98.8\% when compared with the rest of the SOTA implementations.

\end{abstract}


\section{Introduction}
\IEEEPARstart{N}{euromorphic}
computing has garnered a surge in interest, which has led to many innovations with promising results in 
the domain of robotics~\cite{NeuroInRobotics}, healthcare~\cite{NeuroInMedicine}, automotive~\cite{NeuroInVehicle},
agriculture~\cite{NeuroInAgri} and others.
Neuron models were built in multiple forms with varying amount of resemblance to the biological counterpart and is validated by engineering Spiking Neural Networks~(SNN). 
The neuron models such as Izhikevich, and Morris-Lecar are biologically inspired yet are computationally hard to 
emulate when compared to the simpler integrate-and-fire model~\mbox{~\cite{neuroSurvey2017}}.
Every model has unique method of characterizing the performance for a given task, and typically these models are validated for their desired functionality, and performance metrics in software 
framework. 
The proposed Multi-compartment Leaky Integrate and Fire ({\it MCLeaky}) neuron model  belongs to the Integrate and Fire class of neurons. 
The structure of the proposed neuron resembles to the 
'Hippocampus' region of the human brain, and the functional
data flow is established to actively engage in the memory activities.
The proposed neuron model brings in the much needed 
temporal feature extraction for time-series data in the SNN framework, which otherwise requires recurrent neural network's implementation in SNN.

"Hippocampus", structured as a tissue within the parahippocampal gyrus inside the temporal horn, is considered as the memory warehouse of the brain. Interactions occur between this region and the rest of the brain via a structure referred to as entorhinal cortex. Hippocampus is compartmentalized into multiple regions, including 4 cornu-ammonis (CA1 through 4), the dentate gyrus and the subiculum. The flow of information between these compartments takes place via 2 crucial neural circuits referred to as the tri-synaptic and mono-synaptic circuits. The CA3 contains a large recurrent network of axons looping back to the input fibers, which is the main inspiration for the recurrent nature of the {\it MCLeaky} neuron and its internal memristive connections~\cite{Nemati2023}. The "Hippocampus" is responsible for short term memory of human brain and hence, individuals with a well-developed hippocampus have a strong short term memory. CA3 plays an important role in retaining the memory. Hippocampus also decides whether to retain or delete the information when certain parts of the memory is not accessed for adequately long 
duration.

\begin{figure}[htp!] 
\centering
\includegraphics[scale=0.35]{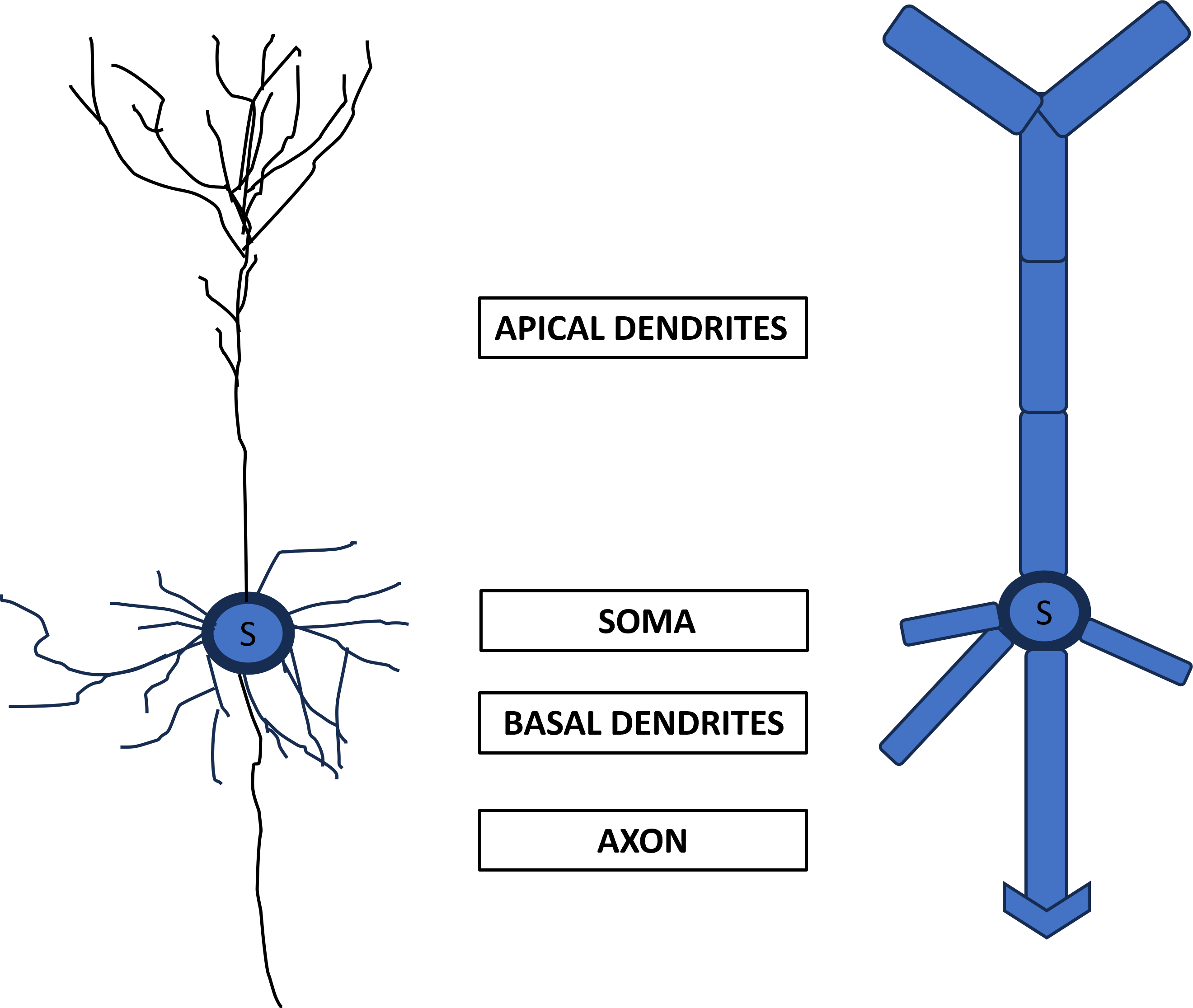}
\captionsetup{justification=centering}
\caption{Schematic representation of the Multi-Compartment Dendritic Neuron model.}
\label{fig:MCModel}
\end{figure}

Similar to Hippocampus functionality,
the proposed {\it MCLeaky} neuron combines two additional
memory components, analogous to CA3 compartment of the human brain.
Additional memory blocks are added with self-looping connections,
forming a CA3 like structure. The result is then supplied to the main memory component.
To realize both looping behavior and the spiking property, the Leaky Integrate and Fire (LIF) neuron forms the most primitive model 
which is widely employed in SNN frameworks.
The Multi compartment is realized as a lumped Resistor and Capacitor model
with finite number of compartments representing ionic equilibrium potentials. 
representing ionic equilibrium potentials. Each of these compartments have similar dynamics and  iso-potential with spatial uniformity in its properties. The dendrites are divided into small compartments and are linked together as shown in the Figrue~\ref{fig:MCModel}.

Long-Short Term Memory units ({\it LSTM}s) are a special form of recurrent neural networks known to perform well with time-series data including speech signal\mbox{~\cite{SpeechSynthesis}}. 
LSTMs also perform well with physiological signals such as electroencephalogram (EEG), electrocardiogram (ECG), photoplethysmogram (PPG), electromyogram (EMG), and electrodermal activity (EDA) and other signals, \hlg{as reported by L. Malviya, et al.~\mbox{\cite{LSTMStressDetection}}}.
They consist of multiple gating mechanism, each performing operations on the input data and on the feedback connections.
\hlg{L. Rezaabad, et al.\mbox{~\cite{LotfiRezaabad2020}}}introduced LSTM method to the neuromorphic computing, with the synaptic current representing the cell state. However, as Spiking LSTMs~(SLSTMs) are designed to replicate the artificial neural network (ANN) counterpart, they lack neurological accuracy and remains a challenge to realize on the hardware. To address this gap, the paper proposes
{\it MCLeaky} neuron to handle time-series data with hardware replicability, besides achieving low power benefits, which is 
thoroughly investigated in this paper.

\hlg{Gao, et al. \mbox{~\cite{Gao2022}} 
discussed }theoretical possibility of implementing dendritic neurons with the aid of MATLAB simulations and FPGA implementation, but 
evaluated for only MNIST datasets.
The test signals were generated directly from ideal voltage 
or current sources, thereby the implemented neuron models 
did not truly indicate the biological representation.
This paper focuses on implementing Multi-Compartment Leaky ({\it MCLeaky}) neuron model in software and the Quantized version of {\it MCLeaky} neurons synthesized on PYNQ-Z1 FPGA, a single board computer based on the Zynq-7020 device from Xilinx.
The proposed neuron model adopted SNN for stress detection 
using physiological EDA signals
was benchmarked against
SOTA Spiking LSTM (SLSTM), and other variants of  SLSTMs.
In SOTA presented by K. Greff, et al.~\mbox{\cite{LSTMComp}}, the authors presented the importance of each gate in LSTM, besides contributing 
towards approximation of the gate operations to realize hardware-friendly variants of SLSTMs.
To summarize, LSTM has 3 main drawbacks for hardware implementation: i) The recurrent connections along the time series create dependencies on the older data, leading to serialization of the inference pipeline, which results in requiring special processing architecture. Most of the current processing accelerators perform well with parallelized vector input, but implementing serialized pipeline leads to processing bottleneck.
ii) LSTM uses multiple non-linear activation functions within each block rather than the more common and simpler ReLU units found in standard ANNs or CNNs, and 
iii) LSTM networks require large memory footprint for storing cell states. This results in a system which is bound by its memory bandwidth.
The proposed {\it MCLeaky} neuron contains compartmental memory block with build in recurrence, alleviating the external memory bandwidth issues, while being modular. It also uses LIF neurons along with a hardware optimized approach and dedicated set of registers to handle time series data.

Quantization-aware {\it MCLeaky} SNNs 
are deduced 
from  Quantization-aware deep evolutionary search runs and are further trained for inference on FPGA. Spiking networks derived from such runs are quantized to form Quantized Deep Spiking Neural Networks~\mbox{({\it QDSNN})} . For testing purposes, 
standard GPU environment for Convolutional ANNs and SNNs, edge devices like Nvidia Jetson Nano$\footnote{NVIDIA Jetson Nano results use Jetson Nano 4GB board running Jetpack SDK 4.6.1 on Ubuntu 18.04LTS operating system and CUDA 10.2. }$, and fixed function hardware of PYNQ-Z1 FPGA boards$\footnote{Xilinx Zynq based FPGA board having 650 MHz dual core Cortex-A9 processor, 53,200 LUTs, 106400 flipflops in the integrated programmable logic (PL) fabric with 143MHz clock and 630 KB fast block RAM, running Theano with Lasagne v0.1. 
}$
were employed. {\it MCLeaky} demonstrated higher level of performance compared to other models with comparable parameter count. Detailed discussion on the performance gain and parameter savings over other implementations is presented in the results section.

\section{EDA Signals and Emotional features} 

Psychological evidence shows that there is a certain correlation between separate emotions~\cite{Lewis2006} including  "Sad" and "Bored", "Happy" and "Amused", which represents a certain degree of specific emotional level. On the other hand,  emotions with the same descriptions 
are distinguished by different order of intensities.
Lang, et al. \mbox{~\cite{Lang1995}}~investigated that emotions are 
measured on valence and arousal components.
Valence component covers the range from being unpleasant 
to a positive emotion, and arousal component covers the range from passive to  a high level of active emotion, which 
relates to humans agitation level. Likewise, stress has both positive and negative interpretations based on its origin. High arousal and negative valence are characteristics of emotional stress, an effective state induced by threatening stimuli as reviewed \hlg{by Christianson, et al.\mbox{~\cite{Christianson1992}}.} 
Stress is considered one of the major cause of concerns in the
urban life, whereby some of the consequences have irreparable
damages~\cite{StressEffects}.
Due to its adverse effects, stress detection and onset of stress
event through real-time analysis, manual assessments, or through wearable sensory systems~\cite{Can2019, Carneiro2019} 
is highly appreciated.

Similar to the other research involving deep neural networks, which require bulk data, the publicly available WESAD dataset is used to train and test the neuron model adopted SNN network for their performance.
Wearable Stress and Affect Detection (WESAD)~\cite{Schmidt2018} dataset consists of human physiological signals derived from a chest-worn device  RespiBAN Professional \footnote{RespiBAN Professional: \url{https://www.pluxbiosignals.com}} and a wrist-worn device Empatica E4 \footnote{Empatica E4: \url{https://www.empatica.com/research/e4/}}. Studies show that among all the physiological signals, electrodermal activity (EDA) is one of the most reliable stress indicators as suggested by studies conducted in ~\mbox{\cite{EDA1, EDA2, EDA3, EDA4, EDA5}}. It measures the change in the activity of the sweat glands’ of a subject, which reflects the changes in arousal. 

An EDA signal has two primary components: the {\em tonic} component that reflects the skin conductance level also denoted as 
SCL. This component measures 
a steadily increasing baseline conductivity that captures 
the long-term, non-specific, and spontaneous fluctuations.
The other \mbox{{\em phasic}} component 
reflects skin conductance response, also denoted as SCR.
This component measures 
a rapidly changing reaction to a specific arousing stimulus, which is evident as pulses or bursts. 
This work primarily focuses on using 
the EDA signal of WESAD dataset acquired from both chest and wrist worn devices. The dataset includes the signals acquired from different  subjects who underwent naturally induced emotions like stress, amusement and baseline along with meditation and recovery states.
The proposed work directly employs the normalized raw signal from the sensor to perform the classification task, thereby the temporal information is kept intact. {\it MCLeaky} neuron employs its memory compartments to handle tonic time series components of the signal.

\begin{figure}[!htp] 
\centering
\includegraphics[scale=0.40]{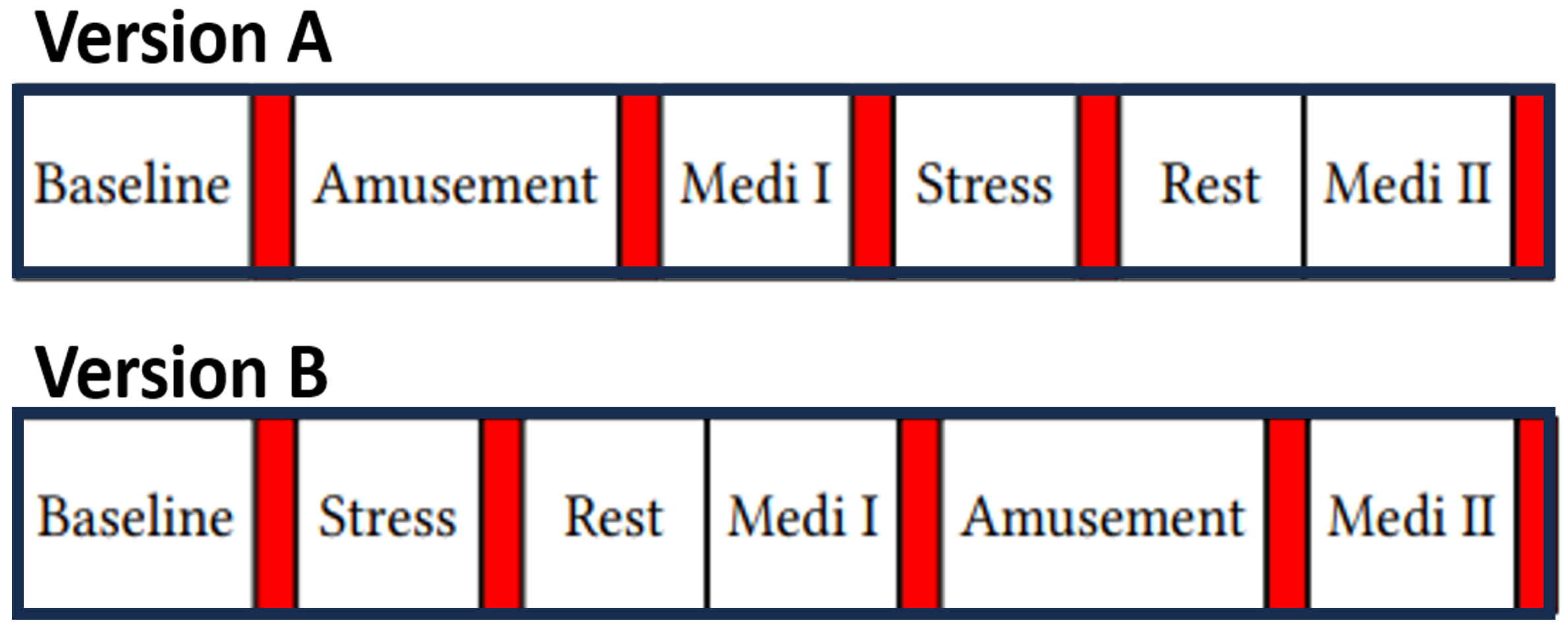}
\caption{The two different versions of WESAD dataset~\mbox{\cite{Schmidt2018}}  study protocol is shown. The red boxes refer to transition state between two emotions where the subjects are 
made to self label the experienced emotion.}
\label{fig:wesadstimuli}
\end{figure}

\begin{figure}[htp!] 
\centering
\includegraphics[scale=0.46]{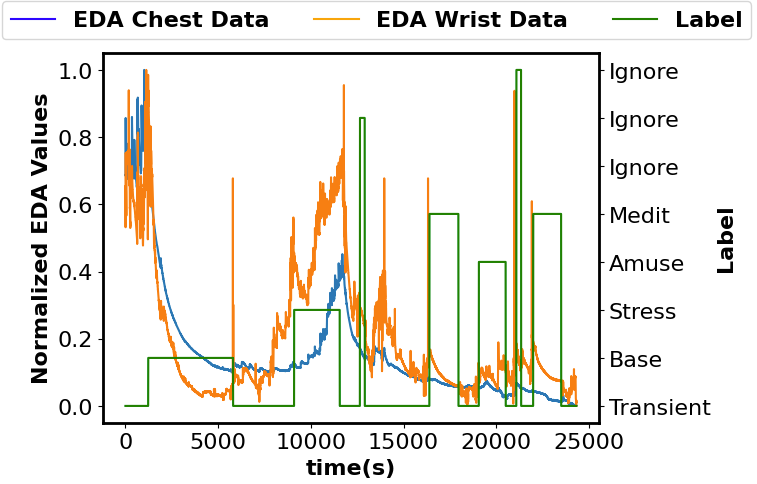}
\caption{A sample of Min-Max normalized EDA chest sensor data~(in blue) and EDA wrist sensor data~(in orange) for subject S2 picked from WESAD dataset 
versus time for the five labels - Amusement~(Amuse), Meditation~(Medit),
Stress, Baseline~(Base), and Transient.
The green line represents the emotion label of that particular sample, which is denoted on the second y axis. Only some of the label definition are provided in WESAD, hence apart from these five, all others are referred to as "ignore" labels . The transient label represents the in-between phase of the recording emotions.}
\label{fig:EDAPreprocess}
\end{figure}

The raw sensor data is min-max normalized and then supplied to the proposed model.
The chest worn RespiBAN device recorded the EDA data at 700~Hz while the wrist worn Empatica E4 recorded the data at 4~Hz. Hence, the data from RespiBAN is downsampled by a factor of 175 to match the lower frequency signal. Hence both high frequency chest signal and low frequency wrist signal 
are available for a given instance.
The data is sliced to bins of equal length. Sliding window  with 50\% overlap is used to perform grouping. The split is also performed such that all the data points in a particular bin belongs to the same label of stress or non-stress. Each resulting bin is then fed to the network for training.
So, the input to the model is a fixed length of one dimensional array containing a chunk of the source signal. 
For the proposed work, we have employed input window size of 256 samples for unimodal method. Unimodal refers to supplying the model with one modality and multimodal refers to supplying multiple modalities in parallel. Input window size for multimodal systems varies with the model-modality combination to obtain the best performance and efficiency.
Figure~\ref{fig:wesadstimuli} summarises the protocol followed in 
WESAD dataset. Here {\it Medi I}, and {\it Medi II} refers to
EDA data acquired while the subjects were in Meditation I, and Meditation II states respectively. The dataset features 2 major stimuli: an amusement condition and a stressful condition. 
Among various states in EDA dataset,
stress state is labelled as it is, while other states are labelled as non-stress as suggested \hlg{by Schmidt, et al.\mbox{\cite{Schmidt2018}}. }
These labelled stress and non-stress EDA data along with 
min-max normalized EDA chest sensor data and EDA wrist sensor data are shown in Figure~\mbox{~\ref{fig:EDAPreprocess}} for a subject randomly picked from WESAD dataset. It is evident from the figure that the samples are unequally distributed and has class annotations for various states (e.g., meditation I, meditation II, recovery, baseline, amusement, and stress).
This work is focused on identifying the 'stress' state (label 2) and therefore the 'non-stress' state is a combination of baseline (label 0) and amusement (label 1) annotations as per the suggestions of WESAD dataset authors~\cite{Schmidt2018}.

For early fusion, Wrist and Chest data bins belonging to the same timestamp are appended to form a double-length one-dimensional tensor array, which we refer to as EDA Combo data. The data is then supplied to the model as input, with a 
train and test split of 80\%:20\% of the dataset. With no other preprocessing steps, the model suits real-time applications. 

There have been similar research work performed with additional preprocessing and feature extraction steps before feeding  to the model, exhibiting the state-of-the-art accuracy~\cite{EDA_Deconv, Ganapathy2020},
but these are time-intensive, thereby affecting the real-time use cases.

To test the generalizability of the {\it MCLeaky} neuron, additional modalities are trained using a MCLeaky based SNN. One of the additional modality includes MIT/BIH's ECG Arrhythmia dataset~\mbox{\cite{MITBIHECG}}. This dataset contains ECG signal recordings from 47 subjects spanning over 24 hours of records. The signal has a sampling frequency of 125Hz. The labels included represent the type of heartbeat, either a normal one or arrhythmic.
\hlg{Studies show that continuous mental stress can cause arrhythmic abnormalities\mbox{~\cite{arrhyDueToStress}}. Hence,
data with normal heartbeat label is considered as non-stress class, and the others are clubbed into the stress class. Stress classification is performed on the resulting dataset. This dataset is also augmented to balance classes, which otherwise results to a skewed dataset.}
A classification task is configured in NAS, similar to the EDA data, to extract the best network model for this dataset. Other modalities tested include the temperature data from WESAD wrist dataset. The temperature signals are extracted from the source files the same way EDA signals are prepared, using a sliding window with 50\% overlap. The tested configuration includes both the unimodal temperature data with 256 samples per window and the multimodal EDA+temperature data with 128 samples of each signal per window.

The results section of this paper also discusses the experiments performed with the in-house sensor data acquired from a subject.
The experiment was performed to validate 
the functionality of the proposed SNN models in real-world scenarios. 
The in-house sensor data was applied to 
validate the performance of the trained model.
Data is acquired from a SeedStudio Grove Galvanic Skin Response (GSR) sensor with the help of an Arduino board Uno R3.
The sensor response is captured at 4~Hz to a local machine and normalized before supplying to the SNN model.

\section{System Design with Stateful Spiking Neurons}

\subsection{{\it MCLeaky} Neuron and Quantization}

\begin{figure*}[htp!] 
\centering
\includegraphics[width=\textwidth]{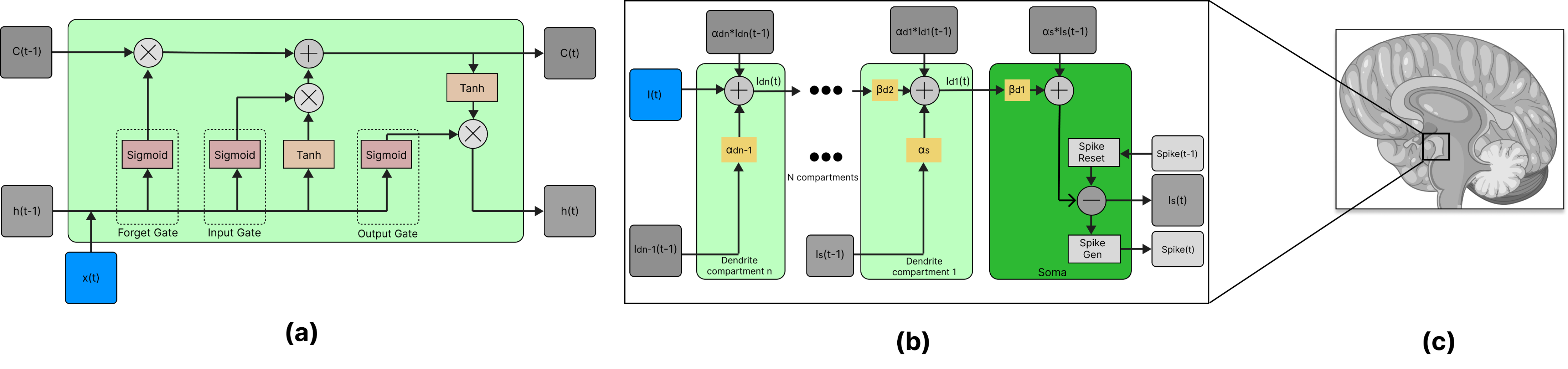} \\
    \vspace{15mm}
\includegraphics[scale=0.35]{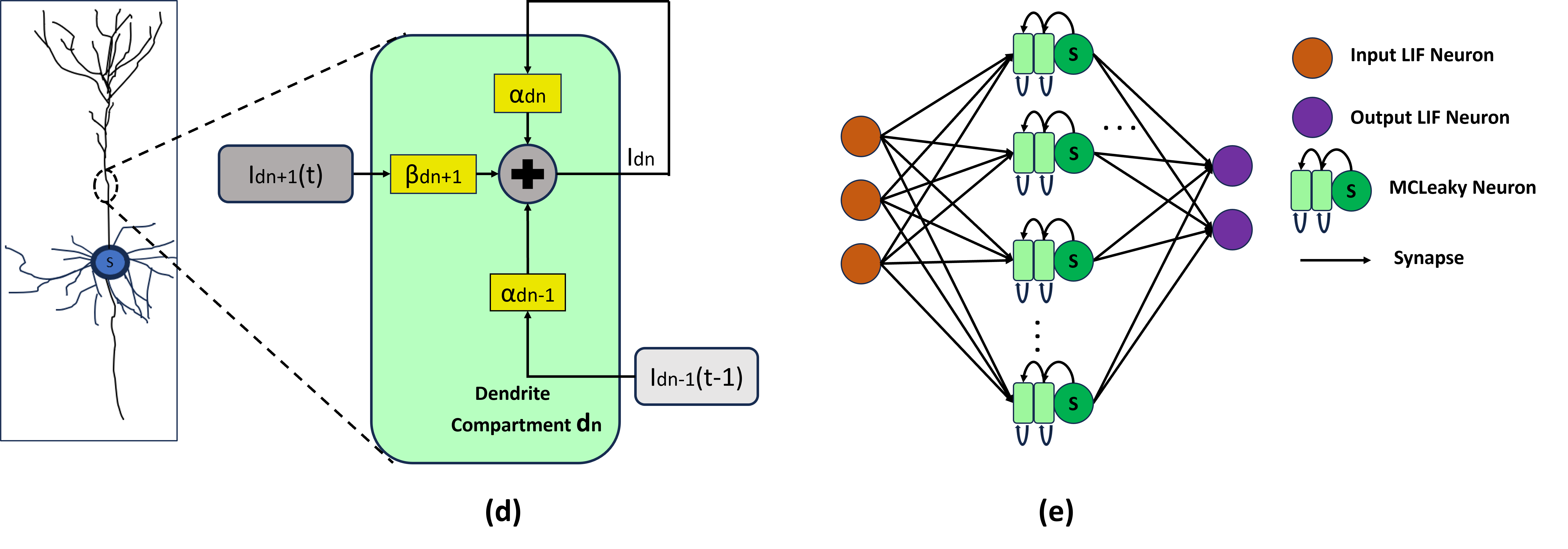}
\caption{Schematic representation of:
(a)~LSTM network, showcasing all gates,
(b)~Proposed {\it MCLeaky} neuron, consisting of multiple dendrite compartments and a single Soma block,
(c)~Highlighted Hippocampus location, that inspired the {\it MCLeaky}'s design,
(d)~Multi-dendritic Compartment neuron model with a Soma and multiple dendrite compartments.
Besides the membrane potential, which portrays  
the memory of the Soma, each 
of the compartment is enabled with
decaying memory in the form of the dendritic potentials and recurrent connections with decay factors. (e)~Illustration of {\it MCLeaky} neuron-based feed-forward SNN with recurrent connections. The built-in memory component in the {\it MCLeaky} neuron, induces sparsity for {\it MC-SNN} model, as compared to
the LIF-based SNNs.}
\label{fig:lstm_mcl}
\end{figure*}

Both LSTMs and {\it MCLeaky} are aimed for real-time data processing and employs storage cells. LSTMs are primarily realized as a mathematical model.
They use non-linear activation functions, thereby 
the time-dependent pipeline incorporated with non-linear function makes then hard to design in hardware.
By design, the building blocks of the {\it MCLeaky} neuron are hardware deployable. 
The mathematical model representing {\it MCLeaky} can be decomposed to multiple 
hardware blocks including 8 bit registers, adders, accumulators, and comparators. Besides {\it MCLeaky} uses LIF as activation function and has a recurrent connection.

All the SNN models were either built on PyTorch or its derivative SNNTorch~\cite{SNNTorch} frameworks. 
For Quantized Deep SNNs (\mbox{\it QDSNN}), the
Brevitas~\mbox{\cite{brevitas}} platform was employed to evaluate the model.
{\it MCLeaky} neuron was initially implemented in software.
Structure of LSTM and N-compartments {\it MCLeaky} neuron is shown in the Figure~\ref{fig:lstm_mcl}~(a) and Figure~\ref{fig:lstm_mcl}~(b) respectively. 
SNNs built from the proposed {\it MCLeaky} neuron with recurrent connections, also referred to as {\it MC-SNN} is illustrated in Figure~\ref{fig:lstm_mcl}~(e). {\it MCLeaky} neuron mirrors a Leaky-Integrate and Fire (LIF) neuron with an additional memory component represented by decaying membrane potential in 
the recurrent connection, which is referred to as a stateful mode of spiking neurons. 
The {\it MCLeaky} neuron is modelled using two dendritic
and a Soma compartment and is further expressed in the Equation~\ref{eq:MCLeaky}, where
$V_{dx}$ denotes the membrane potential of dendrite compartment $x$, $ V $ denotes soma compartment membrane potential at time $t$, $ \beta $ and $\alpha$ denotes the decay parameter for the membrane potentials derived from the previous stage and its recurrent connections from following stage respectively,
$ I $ denotes input current to the dendritic compartment of {\it MCLeaky} neuron, $reset$, and $\vartheta$ represent the reset spike and threshold values of neuron. Input to the neuron is supplied to one of the compartments with the previously decayed dendritic membrane state. The compound information then flows to additional dendrite components and ends up in the Soma memory. When the Soma membrane potential $v^{t}$ exceeds a certain threshold $\vartheta$, the neuron will elicit a spike, and subsequently, the membrane potential $V^{t}$ is reset.

\begin{equation}
\begin{split}
\label{eq:MCLeaky}
V_{d2}^{t} &= (\alpha_{d2} \times V_{d2}^{t-1})  + 
(\alpha_{d1} \times V_{d1}^{t-1}) + I^{t}  \\
V_{d1}^{t} &= (\alpha_{d1} \times V_{d1}^{t-1}) +
(\beta_{d2} \times V_{d2}^{t}) +  (\alpha_{s} \times V_{s}^{t-1})  \\
V^{t} &= (\alpha_{s} \times V_{s}^{t-1}) + (\beta_{d1} \times V_{d1}^{t}) - (reset \times \vartheta) \\
S^{t} &= SpikeGen(V^{t} - \vartheta)\\
\end{split}
\end{equation}

{\it MCLeaky} based SNN models are trained using Backward Propagation through time (BPTT), where all the spikes generated for a data sample are back-propagated 
post a forward pass run. 
The result of the model is further applied to a count based 
Mean Squared Error~(MSE) loss function.
This count based loss functions take the output spikes and tunes the model to fire 80~\% for all the time-steps related to the correct class and 20~\% for the incorrect class. A spike rate of 80\% for the true positive case is chosen such that the neuron representing the correct class will fire enough times to be classified while also being power efficient. On the other hand, having spike rate of 20\% for the incorrect class ensures at least some information propagation, alleviating any neuron activity suppression.
Quantized
{\it MCLeaky} neuron involves quantization of the state parameters, including  $\alpha$ decay, $ \beta $  decay and threshold values. 
To simplify the quantization process of the {\it MCLeaky} neuron, $ \beta $ and $\alpha$ values were set to a constant value of 0.875, 
and threshold was configured to a quantized 8-bits fixed-point 
format. The $ \beta $ value 
of 0.875 was deduced by 
expanding the decaying exponential value of 2 from Taylor series representation.
The \textit{Q-MCLeaky} neuron, when adopted for Quantized fully connected layer, forms the required Quantized-Deep-learning-SNN, which is denoted as \textit{QDSNN}.
The quantized, fully connected layer is represented entirely by quantized weights and biases which forms the  state parameters.

\subsection{Decaying Memory LSTM and Simplified
Spiking LSTMs}
CNNs applied on time-series signals are typically accompanied by a standard fully connected layer,
 or an LSTM layer followed by activation. Similarly, in this work 
 {\it MCLeaky} neuron based
 SNNs are constructed with a fully connected layer, or an {\it SLSTM} layer followed by an activation layer.
 This work also establishes the SNNs with 
 standard LIF and current-based LIF (C-LIF) neurons to benchmark the impact of {\it MCLeaky} neuron on SNNs. 
 The corresponding SLSTM network for spiking inputs and its variants are also established to benchmark the proposed work.
 To mimic the forgetfulness of the human brain, a decaying unit is added to the memory component of standard {\it SLSTM}
 which models memory loss at a specific rate.
 The novel variant of decaying memory devised SLSTM ({\it DM-SLSTM}) is modelled as stated in the Equation~\ref{eq:DM-SLSTM}, 
 followed by the Equation~\ref{eq:SLSTM} representing {\it SLSTM} model.

\begin{equation}
\begin{split}
\label{eq:DM-SLSTM}
    &x_t = \alpha \times x_{t-1} + I_t \\
\end{split}
\end{equation}

For additional benchmarking, {\it SLSTM} models are further simplified to concede fewer parameters, yet attain comparable accuracy. Three simple variants of \textit{SLSTMs} referred to as \#1, \#2 and \#3 along with the standard \textit{SLSTM} and the newly formulated  \textit{DM-LSTM} are setup and compared with \textit{MCLeaky-SNN} models.
The set of expressions listed in Equation~\ref{eq:SLSTM} represents the standard {\it SLSTM} neuron. Here, 
$ \sigma $ denotes the Sigmoid operation and $\odot$ represents hadamard product. $x_t$ represents input to the layer, $mem_t$ represents the memory component at time $t$, $W_{ab}$ and $b_{ab}$ represent weights and bias for that particular data matrix, 
where $a$ is applicable for any inputs $i$,
$h$ is for hidden, and $b$ is one of the other input~$i$,  with forget($f$), output($o$) and gate($g$) functions also specified. 
$i_t$, $f_t$, $o_t$, $g_t$ represent the output of the gate equations. $I_t$ is the output of the entire {\it SLSTM} cell, 
which is further normalized by transforming to $tanh$ function as listed in the last expression 
of Equation~\ref{eq:SLSTM}.
The other three simple variants of {\it SLSTM} are formed by eliminating the gate function component, 
as stated in the Equation~\ref{eq:SLSTM}. 
Equation~\ref{eq:SLSTM1} refers to the first variant of simplified {\it SLSTM}~(\#1), by skipping input and its weight matrices. 
Equation~\ref{eq:SLSTM2} represents the second variant of simplified {\it SLSTM}~(\#2), which is deduced by skipping the 
bias terms from the first variant of Simplified {\it SLSTM}~(\#1).
Equation~\ref{eq:SLSTM3} models the third variant of 
simplified {\it SLSTM}, keeping only the bias terms in the gate function.
Although no information flows through the gates in these cases, the remaining parameters are adaptively trained during back-propagation, which aids in evolving a fine-tuned model
for the unseen input data.

\begin{equation}
\left.
\begin{split}
\label{eq:SLSTM}
    &i_t = \sigma(W_{ii} x_t + b_{ii} + W_{hi} mem_{t-1} + b_{hi}) \\
    &f_t = \sigma(W_{if} x_t + b_{if} + W_{hf} mem_{t-1} + b_{hf}) \\
    &o_t = \sigma(W_{io} x_t + b_{io} + W_{ho} mem_{t-1} + b_{ho}) \\
    &g_t = \tanh(W_{ig} x_t + b_{ig} + W_{hg} mem_{t-1} + b_{hg}) \\
    &I_t = f_t \odot  I_{t-1} + i_t \odot  g_t \\
    &mem_t = o_t \odot  \tanh(I_t) \\
\end{split}
\right\}
\end{equation}

\begin{equation}
\left.
\begin{split}
\label{eq:SLSTM1}
    &i_t = \sigma( b_{ii} + W_{hi} mem_{t-1} + b_{hi}) \\
    &f_t = \sigma(b_{if} + W_{hf} mem_{t-1} + b_{hf}) \\
    &o_t = \sigma( b_{io} + W_{ho} mem_{t-1} + b_{ho}) \\
\end{split}
\right\}
\end{equation}

\begin{equation}
\left.
\begin{split}
\label{eq:SLSTM2}
    &i_t = \sigma( W_{hi} mem_{t-1}) \\
    &f_t = \sigma( W_{hf} mem_{t-1}) \\
    &o_t = \sigma( W_{ho} mem_{t-1}) \\
\end{split}
\right\}
\end{equation}

\begin{equation}
\left.
\begin{split}
\label{eq:SLSTM3}
    &i_t = \sigma(b_{hi}) \\
    &f_t = \sigma(b_{hf}) \\
    &o_t = \sigma(b_{ho}) \\
\end{split}
\right\}
\end{equation}

\subsection{Neuro-Evolution of QDSNNs (NE-QDSNN)}
Rather than iterating through various hyper-parameters and layer combinations to derive the optimal network for a specific neuron component or layer, the Neural Architecture Search (NAS) paradigm was employed for this work. 
The neuro-evolutionary (NE) framework is built on Genetic algorithms (GA) and Dynamic Structured Grammatical Evolution (DSGE), enabled with network hyperparameters. 
GA-level represents the sequence of evolutionary units to formulate \textit{QDSNN} encoded in macro-structures. Each unit in the sequence is a non-terminal symbol representing a valid DSGE-level genotype. The genotype comprises of the SNN layers, including dropout, the order of these layers, the learning-rate and methods, and the classification method. DSGE-level specifies the parameters used at the GA-level. The range of parameters and possibilities are coded in a backus-Naur form (BNF) grammar. Beginning from the Non-terminal unit of GA-level, DSGE allows expanding the set of individual genotypes until all symbols in the phenotype are non-terminals. By changing the BNF grammar, different bit width quantization networks with \textit{MCLeaky} or LIF spiking neurons along with different neuron parameters are generated.

To improve energy efficiency and achieve richer neuron dynamics, the threshold parameter of {\it MCLeaky} neurons, {\it DM-LSTMs}, and simplified {\it SLSTMs} are made learnable. Additionally, decay parameters of the dendritic membrane potentials connected recurrently are also made trainable.
The threshold and decay parameter for each neuron is learnt during the training phase of the model. The recurrent connections for the decaying memory of the neuron enhance the dynamics, leading to the possibility of reduced computations and thereby achieving power efficiency. The experiments are performed using a modified version of the ANN NAS tool set detailed in Assuncao F., et al.\mbox{~\cite{DENSER}}.
In addition to the NAS framework, various spiking neurons and layers, Spike-based activations, loss functions,
{\it DM-SLSTM}, simplified {\it SLSTMs}, and fast-sigmoid surrogate gradient in Backward Propagation through time (BPTT) 
are included to prioritize spiking output ratio and power efficiency.
Packages employed include SNNTorch for SNNs and Brevitas for Qunatized SNNs. Individual models are trained using BPTT and configured with cross entropy (CE) rate loss. SNNs require the data to be presented as spikes. To convert floating point values to spikes, rate based encoding as proposed \hlg{by Edgar D et al.~\mbox{\cite{rateEncding}} is performed}. Spikes are generated from a binomial distribution with mean as the input signal samples. Hence, each signal floating point value is converted to a fixed length 1D array containing Ones~(spike) and Zeros~(no spike). This results in converting a signal data point array to a 2D matrix where a row represents a data point and a column represents time axis~\mbox{\cite{spikeCodingReview}} .  Accuracy is used as the fitness metric 
to find the best individual model from the evolutionary search runs. \footnote{NE-QDSNN evolutionary search runs use NVIDIA RTX A4000 GPU with CUDA v12.0.} 

\begin{figure*}[htp!] 
\centering
\includegraphics[scale=0.45]{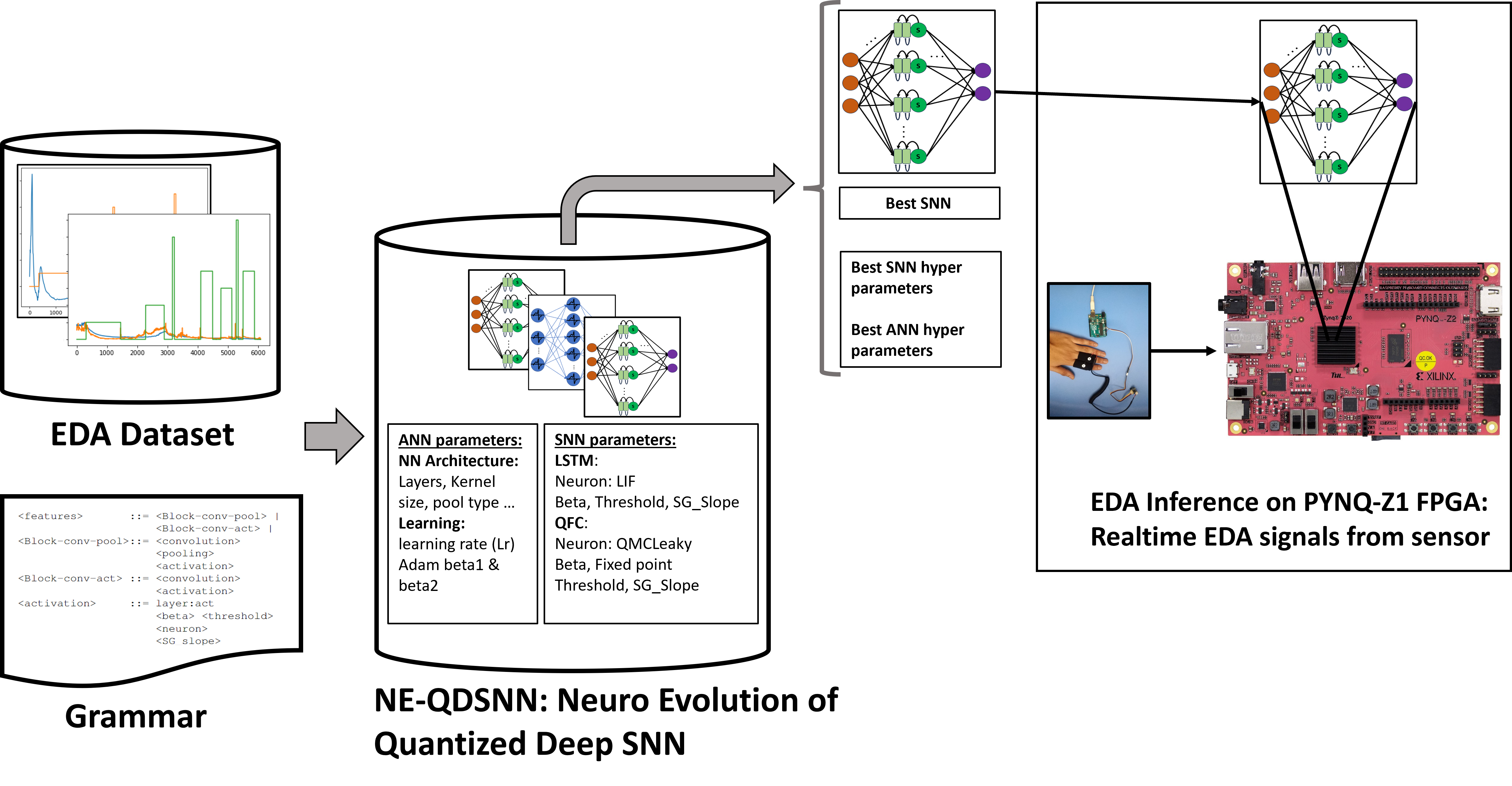}
\caption{System design approach: EDA signals from WESAD dataset are rate encoded and fed to the proposed {\it NE-QDSNN} along with the pre-defined Grammar. {\it NE-QDSNN} generates best choices of {\it QDSNN} along with stateful spiking neuron parameters. {\it QDSNN} model is further validated for an in-house data-acquisition unit built for EDA signal.}
\label{fig:NEQDSNN}
\end{figure*}

As quantization is important for achieving real-time inference on edge computing devices, Hardware-quantized Neural Architecture Search is important in generating cost-effective SNNs. This work demonstrates the effectiveness of hardware-aware {\it QDSNN} evolved from Quantized-NAS, which is enabled to train the networks on physiological EDA signals and equipped with stateful spiking neurons for generating a power-efficient stress detection model 
as depicted in Figure~\ref{fig:NEQDSNN}.
Results from live data were acquired by training the best model extracted from the NAS framework run separately, and further generating the inference on the trained model.
{\it NE-QDSNN} approach is employed for the MIT-BIH ECG dataset to analyse the robustness of the {\it MCLeaky} neurons, and the best model is extracted from the NAS framework for stress detection.
All the hyper-parameters and parameters employed for training the model for stress detection are summarised in Table~\ref{table:EvoParam}.

\begin{table}[!htp]
         \begin{center}
            \caption{Evolutionary Parameters and Hyper-parameters for establishing {\it QMCLeaky - QDSNN}.}
            \resizebox{0.48\textwidth}{!}{
        \begin{tabular}{|c|c|}
        \hline
         \textbf{Evolutionary parameters} & \textbf{Value} \\
         \hline
         Number of Generations &  16     \\
         \hline
         Number of Parents & 16  \\
         \hline
         Number of Offsprings & 16  \\
         \hline 
         Add layer rate & 0.15  \\
         \hline 
         Reuse layer rate & 0.15  \\
         \hline 
         Remove layer rate & 0.25  \\
         \hline 
         DSGE Layer rate & 0.15  \\
         \hline 
         Macro Layer rate & 0.3  \\
         \hline 
         Fitness function & Accuracy  \\

         \hline 
         \addlinespace 
         \hline 
         
         \textbf{Training Parameters} & \textbf{Value}  \\
         \hline 
         Number of epochs & 3  \\
         \hline 
         Batch size & 24  \\
         \hline 
         Encoding & Rate encoding  \\

         \hline          
 
         Loss function & Cross entropy rate loss   \\
 
         \hline 
         
         Optimizer & Adam  \\
         \hline 
         
        \end{tabular}
}
         \label{table:EvoParam}
         \end{center}
\end{table}

A glimpse of the grammar employed for defining the search space 
is illustrated in Figure~
\ref{fig:grm}. It uses a custom syntax to define two levels of structure: macro and micro levels. The micro level includes individual parameter definitions such as decay rate, input and output shapes, quantization bit length, dropout rate, $\alpha$ and $\beta$ decay parameters and threshold on which evolutionary mutation operations are applied. The syntax defines the variable name, variable datatype, number of values it holds, minimum and maximum among them. Considering $<q\_bits>$ as referred from Figure~\ref{fig:grm}, 
it stores one integer, and ranges from 4 to 32, inclusive. Additional definitions include layer specifications such as
$<fully-last>$ stating the type of the layer and the 
number of units in the same line.
The macro level combines multiple micro level definitions to form
full-fledged block of network 
on which crossover operation is performed. 
They include $<classification>$, $<fully-connected>$ and $<output>$. 
$<output>$ represents a $<fully-last>$ layer followed by $<activation-final>$ activation layer.

\begin{figure}[!htp]
    \centering
    \includegraphics[width=\linewidth]{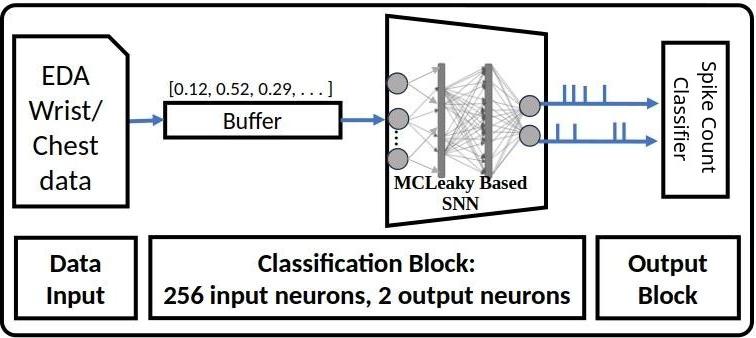}
    \caption{Proposed inference pipeline of an {\it MCLeaky} based Spiking Neural Network. The pipeline does not use pre-processing step, which makes it a real-time inference engine.}
    \label{fig:mclPipeline}
\end{figure}

The resulting model from the NAS runs is extracted to perform inference on the taregted class. The schematic of the inference pipeline designed for {\it MCLeaky} based SNNs is 
depicted in the Figure~\mbox{\ref{fig:mclPipeline}}.
 The block diagram shows a 3 stage pipeline architecture with input data supplied to SNN classification block, resulting in output.
 As depicted, {\it MCLeaky} incorporated SNN skips the traditional pre-processing steps, 
 resulting in real-time use-case with 3 stage inference pipeline.
 
\begin{figure}
\begin{small}
\begin{verbatim}
<features> ::= <fully-connected>
               <activation_cl>
<beta>     ::= [beta,float,1,0,1]
<threshold>::= [threshold,float,1,0.1,10]
<sur_grad_slope>::=[grad_slope,int,1,1,30]
<classification> ::=<fully-connected>
                    <activation_cl> 
                   |<dropout>
<dropout>::=layer:dropout[rate,float,1,0,0.5]
<fully-connected>::=layer:qfc 
                    [num-units,int,1,16,1024]
                    <q_bits>
<output> ::= <fully-last> <activation_final>
<activation_cl>::=layer:act <beta><threshold>
                  <neuron_cl><sur_grad_slope>
<neuron_cl>       ::=neuron:QMCLeaky
<activation_final>::=layer:act <beta>
                     <threshold> <neuron_cl>
                     <sur_grad_slope>
<fully-last>::=layer:qfc num-units:2 <q_bits>
<q_bits>   ::= [q_bits,int,1,4,32]
<learning> ::= <adam> 
<adam>     ::= learning:adam
               [lr,float,1,0.001,0.1]
\end{verbatim}
\end{small}
\caption{Grammar Configuration for {\it QMCLeaky} model.}
\label{fig:grm}
\end{figure}

\section{Multi-Compartment (MC) Spike Processor}

\begin{figure}[htp!] 
\centering
\includegraphics[scale=0.21]{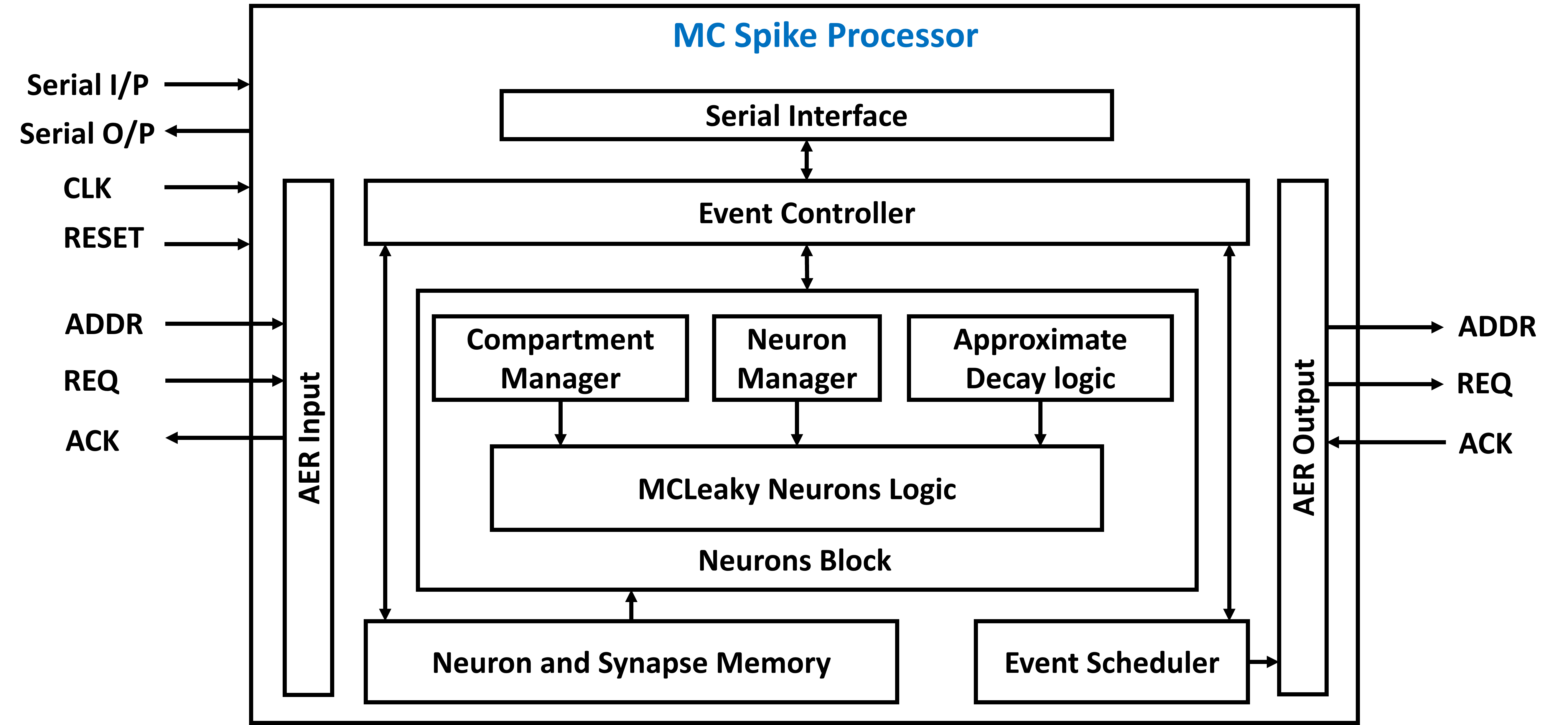}
\caption{Hardware Architecture of Multi-Compartment ~(MC) Spike processor, which is deployed on PYNQ-Z1 FPGA.}
\label{fig:HWMCSpikeProc}
\end{figure}

The implementation of the proposed {\it MCLeaky} based {\it QDSNN} model on PYNQ-Z1 FPGA is depicted in Figure~\ref{fig:HWMCSpikeProc}.
The Event Controller, the communication interfaces and Scheduler FIFOs are inspired from Online-Learning Digital Spiking Neuromorphic (ODIN) Processor architecture published in Frenkel C., et al.~\cite{Frenkel2018}.
All the {\it MCLeaky} neuron parameters and synapse states are stored in SRAMs, but it maps to BRAM when deployed  on FPGA.
Address Event Representation~(AER) is an asynchronous protocol that handles real-time input spikes or events and generates an output spike with the neuron ID and the corresponding time-step. Scheduler FIFOs are implemented as buffers for the AER transactions.

\subsection{Event Controller and Scheduler}
The individual states of $N$ neurons and $N^2$ synapses are stored in Neuron SRAM and Synapse SRAM memory; a Spike Manager handles the neuron and synapse updates to emulate $N \times N$ crossbar.
The Event controller decodes the AER packets in the FIFO received from the input AER and checks if the input time-step is equal to the current time-step. 
If so,
the integration procedure for this time-step is not completed yet, and 
post neuron transformation, voltage continues to integrate.
Otherwise, it marks the end of the current time-step so that neuron triggers the spike. 
The scheduler handles the spike events from all neurons and arbitrates between external and internally generated neuron events. Spiking {\it MCLeaky} neurons are set to send 14-bit event packets to the scheduler. The packets contain 8-bit address of the source neuron, a 3-bit field indicating the number of spikes to be generated minus one and a 3-bit field quantifying the Inter-Spike-Interval. The functionality of the event controller, the event scheduler and input/output (IO) FIFOs are elaborated \hlg{by Frenkel, et al.\mbox{~\cite{Frenkel2018}}}. The initial threshold of the neurons and the offline trained weights are pre-loaded to synaptic RAM through
the SPI interface.

\subsection{{\it MCLeaky} Neurons Block}
Neuron Manager controls each of the 256 neurons and configures
either {\it MCLeaky} or {\it LIF} neuron that is under evaluation.
The former is proposed, whereas the latter is included for benchmarking purposes.
{\it MCLeaky} includes a single Soma and parametric dendrite compartments that are connected together. 
The neuron manager coordinates between the Compartment Manager and the neurons to program the initial thresholds. The compartment manager configures the number of compartments in each of the {\it MCLeaky} neurons. The spike events determine the reset
status of the neuron.  The Event controller decodes the AER packets,  integrates voltage, and elicit a spike upon reaching the threshold.
Discrete-time defined {\it MCLeaky} neuron 
representing Multi-Compartment logic 
is expressed 
in Equation~\mbox{~\ref{eq:MCLeaky}}.
{\it MCLeaky} neurons and Synaptic weights are 8-bit quantized, both for training and inference. As the neurons are time-multiplexed, their states and parameters are stored in a standard single-port SRAM and pre-loaded from SPI interface before inference.

\begin{figure}[htp!] 
\centering
\includegraphics[scale=0.25]{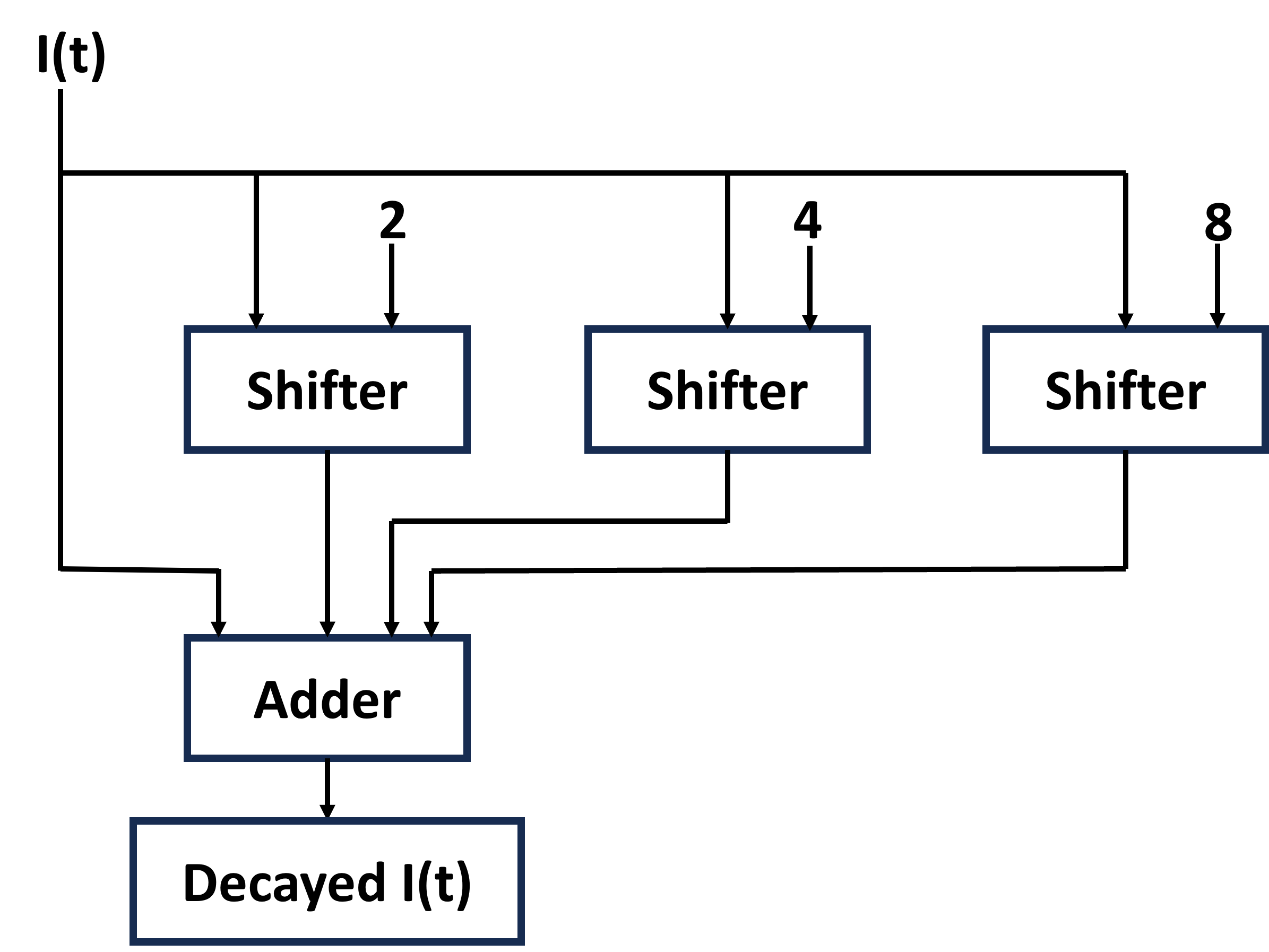}
\caption{Taylor series of exponential decaying dendrite or soma current, which is implemented on
PYNQ-Z1 FPGA.}
\label{fig:HWDecayedCur}
\end{figure}

\begin{figure}[htp!] 
\centering
\includegraphics[scale=0.25]{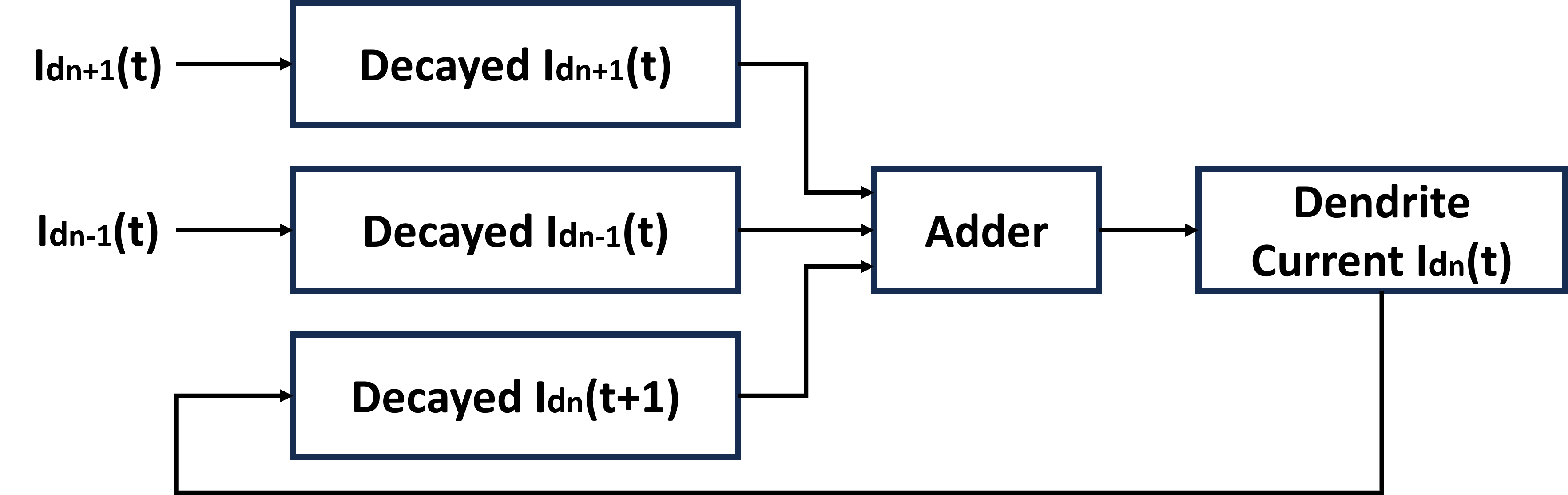}
\caption{Hardware architecture of dendrite block, that is designed for 
PYNQ-Z1 FPGA.}
\label{fig:HwDendrite}
\end{figure}

\begin{figure}[htp!] 
\centering
\includegraphics[scale=0.25]{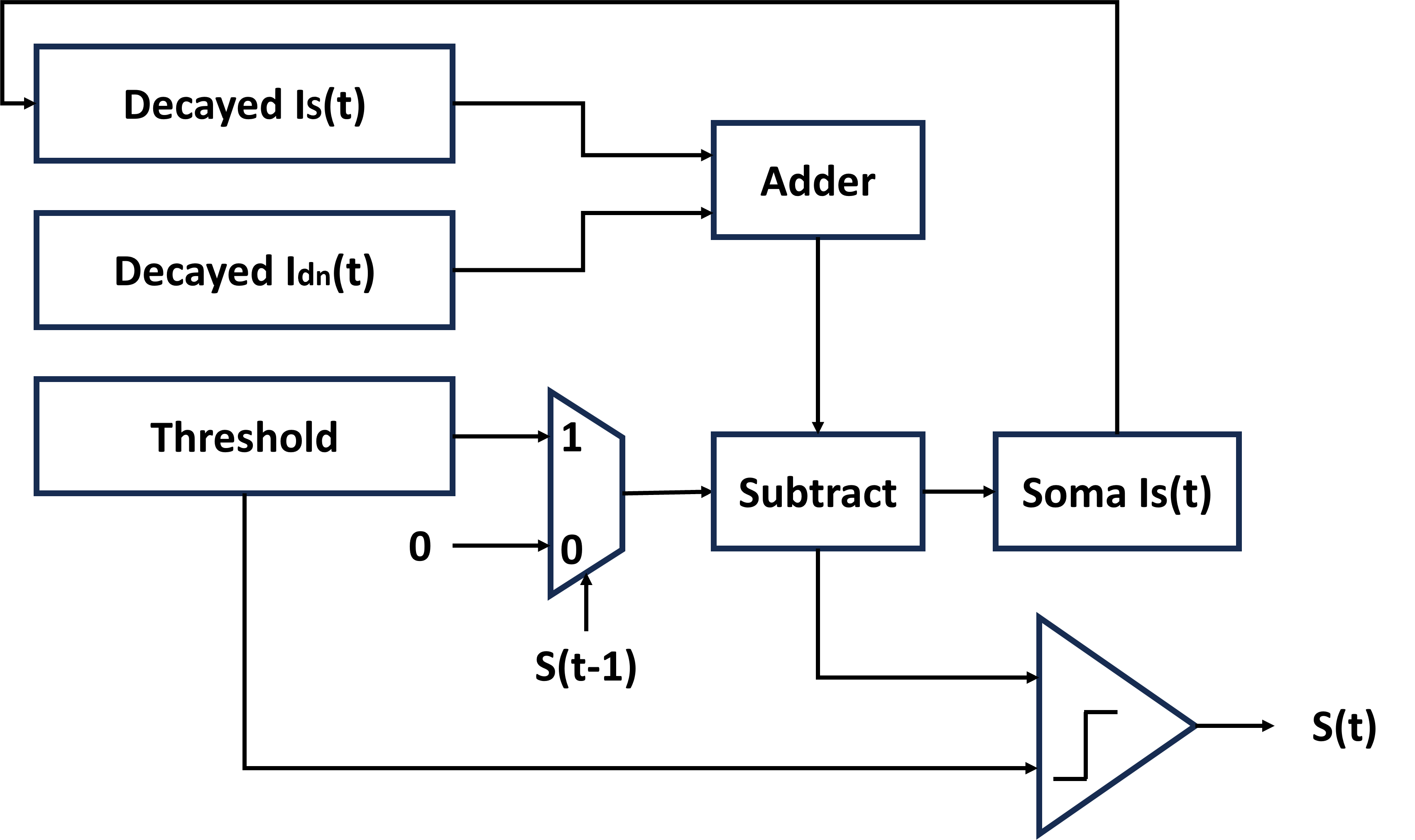}
\caption{Hardware Architecture of Soma block. Design implemented on PYNQ-Z1 FPGA.}
\label{fig:HWSoma}
\end{figure}

The delay logic offers a decay value of $0.875$ to each neuron as a first step in the Inference model. The neuron's membrane potential gets accumulated, with the incoming spikes multiplied by weights till it reaches the threshold, post which the membrane potential is reset. The Approximate decay logic is formulated with the Taylor series expansion value of the $e^{(-t/\tau)}$ for three terms. $0.875$ 
is considered to be an optimal decay constant with $\tau=2$ for the {\it QMCLeaky} neurons. 
The Approximate logic, Dendrite block and Soma block microarchitecture are shown in Figures~\ref{fig:HWDecayedCur}, \ref{fig:HwDendrite},  and \ref{fig:HWSoma} respectively. The proposed {\it QMCLeaky-QDSNN} is also evaluated for MIT-BIH ECG dataset~\cite{MITBIHECG}. FPGA implementations are summarized in Table \ref{table:FPGAResults}. The proposed {\it QMCLeaky-QDSNN} further reduced the LUTs and Flops utilization by 4.6~$\times$ and 2.4~$\times$ compared with the most compact design stated in~\cite{Frenkel2018} and has no Digital Signal Processing (DSP) slices, thereby drastically reducing the power consumption.

\begin{table}[!htp]
    \centering
    \caption{Comparison of FPGA Implementation Results with existing SNN hardware.}
    \label{table:FPGAResults}

       \resizebox{0.48\textwidth}{!}{
    \begin{tabular}{|c|c|c|c|}
    \hline
        \textbf{Design}  & Frenkel\textbf{~\cite{Frenkel2018}} & Wang\textbf{~\cite{Wang2017}} & \textbf{Proposed {\it QMCLeaky-QDSNN}} \\
        \hline

        Accuracy  & 84.5\% & 89.1\% & 92.21\% \\
        \hline
        FPGA Platform & Zynq 7020 & Virtex-6 & Zynq 7020  \\
        \hline                            
        Number of Neurons & 256 & 1591 & 256  \\
        \hline                
        Number of Synapses & 65536 & 638208 & 65536 \\
        \hline        
        LUT Utilization & 5114 & 71666 & 1123  \\
        \hline        
        FF Utilization & 4248 & 50921 & 1745  \\
        \hline                
        Block RAM Utilization & 2.3\%  & - & 2.8\%  \\
        \hline                
        DSP Utilization & 0\% & - & 0\%  \\
        \hline                
     \end{tabular}
     }

 \end{table}


\section{Results}

\begin{figure*}[htp]
    \centering
    \includegraphics[width=\textwidth]{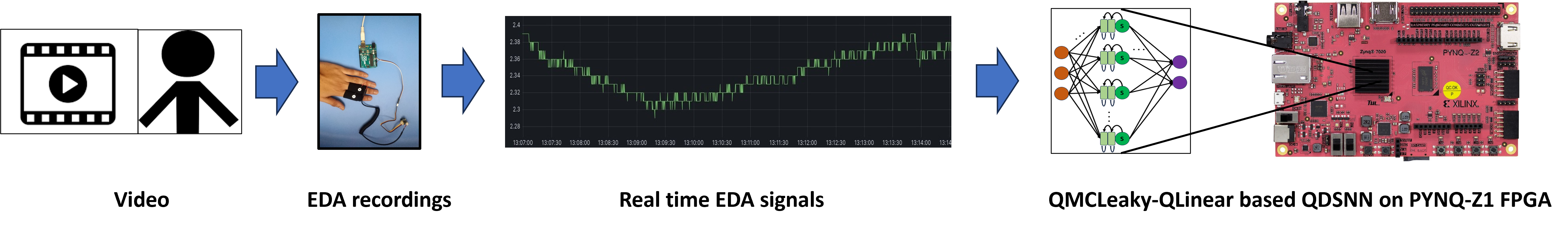}
    \captionsetup{justification=centering}
    \caption{Schematic representation of {\it QDSNNs} Inference pipeline for stress detection 
using chaotic electrodermal states for an edge computing scenario.
}
    \label{fig:EmotionPipeline}
\end{figure*}

\begin{table*}
    \centering
    \caption{Comparison of the proposed work with state-of-the-art~(SOTA) published results for stress detection with EDA signals.}
    \label{tab:paperComparison}
    \begin{tabular}{|c|c|c|c|c|}
        \hline
        \textbf{Source} & \textbf{Model} & \textbf{Data} & \textbf{Test Accuracy} & \textbf{Input Size}\\
        \hline
        M.A. Fauzi et al. ~\cite{ali2022} & Statistical Features + Linear Regressor & Multimodal with EDA & 85.75  & 240 \\ 
        \hline
        M. Quadrini ~\cite{quadrini2022} & Angular fields + CNN & Multimodal with EDA & 91.1 & 293 \\
        \hline
        Z. Stržinar et al. ~\cite{ZigaS2023} & Statistical Features + ANN & EDA Wrist & 84 & 240  \\
        \hline
        L. Huynh et al. ~\cite{CNN_NAS_EDA} & Filters + NAS CNN & EDA Wrist & 79.24 & 240 \\
        \hline
        \textbf{Proposed CNN} & Raw Signal + CNN & EDA Combo & 91.8 & 128+128 \\
        \hline
        \textbf{Proposed MCLeaky-SNN} & Raw Signal + MCLeaky SNN & EDA Combo & 98.8 & 128+128\\
        \hline
        \textbf{Proposed QMCLeaky-SNN} & Raw Signal + QMCLeaky SNN & EDA Combo & 91.84 & 128+128\\
        \hline
        \textbf{Proposed DM-SLSTM SNN} & Raw Signal + DM-SLSTM SNN & EDA Combo & 89.56  & 128+128\\
        \hline
        \textbf{Proposed CNN} & Raw Signal + CNN & Realtime EDA Data & 87.5 & 256 \\
        \hline
        \textbf{Proposed MCLeaky SNN} & Raw Signal + MCLeaky SNN & Realtime EDA Data  & 93.75 & 256 \\
        \hline
        \textbf{Proposed DM-SLSTM SNN} & Raw Signal + DM-SLSTM SNN & Realtime EDA Data  & 93.75 & 256 \\
        \hline
        \textbf{Proposed MCLeaky-SNN} & Raw Signal + MCLeaky SNN & MIT-BIH ECG  & 93.47 & 247 \\
        \hline
        \textbf{Proposed QMCLeaky SNN} & Raw Signal + QMCLeaky QDSNN & MIT-BIH ECG  & 92.21 & 247 \\
        \hline
        \textbf{Proposed MCLeaky SNN} & Raw Signal + MCLeaky SNN & EDA Wrist & 80.77 & 256 \\
        \hline
        \textbf{Proposed MCLeaky SNN} & Raw Signal + MCLeaky SNN & EDA Chest & 86.15 & 256 \\
        \hline
        \textbf{Proposed MCLeaky SNN} & Raw Signal + MCLeaky SNN & Temperature Wrist & 90 & 256 \\
        \hline
        \textbf{Proposed MCLeaky SNN} & Raw Signal + MCLeaky SNN & EDA + Temp Wrist & 91.79 & 128+128 \\
        \hline
        \textbf{Proposed MCLeaky SNN} & Raw Signal + MCLeaky SNN & EDA Combo + Temp Wrist & 99.70 & 48+48+48 \\
        \hline
    \end{tabular}
\end{table*}

The results listed here are the models extracted from the NAS runs, and hence 
no two models evolved from NAS, either quantized or non-quantized, will necessarily be similar.
This ensures that the best-performing model with richer neuron dynamics is compared across the benchmarks, which also implies that
a fixed model architecture to fit all the neurons does not exist.
NAS search space parameters are listed in Table~\ref{table:EvoParam}.

Table~\ref{tab:paperComparison} compares primary models introduced in this paper against models from ~\cite{ali2022,quadrini2022,ZigaS2023,CNN_NAS_EDA}. Results included in this table are also explained in depth in later parts of this section.
Initial investigations include 256 samples of EDA Chest and Wrist data from the WESAD dataset appended as mentioned above. {\it SLSTM} configurations investigated in this work include
the approximate variants~(\#1, \#2, and \#3), the simpler {\it SLSTMs} and {\it DM-SLSTM}.
Comparison among these {\it SLSTM} configurations is listed
in Table~\ref{tab:SLSTMComp}. Regarding classification accuracy, {\it DM-SLSTM} exhibited the best over others. However, it concedes the highest parameter count, which 
is attributed to the additional trainable decay and offset parameters. The parameter count and model performance follow the order of gate 
function complexity.

\begin{table}[!htp]
    \centering
    \caption{Comparison of various LIF based Spiking LSTM networks.}
    \label{tab:SLSTMComp}
    \begin{tabular}{|c|c|c|}
        \hline
        \textbf{SNN Configurations} & \textbf{Accuracy (\%)} & \textbf{Parameter Count} \\
        \hline
        {\it DM-SLSTM} & 89.56 & $3M$ \\ 
        \hline
        {\it SLSTM} & 88.78 & $1M$ \\
        \hline
        SLSTM \#1 & 86.73 & $700K$ \\
        \hline
        SLSTM \#2 & 70.41 & $600K$ \\
        \hline
        SLSTM \#3 & 70.41 & $600K$ \\
        \hline
    \end{tabular}
\end{table}

\begin{table}[!htp]
    \centering
    \caption{Comparison of EDA combo Test Accuracy and model parameters for different SNN models.}
    \resizebox{0.48\textwidth}{!}{

    \begin{tabular}{|c|c|c|c|}
    \hline
        \textbf{Network} & \textbf{Test Accuracy in \%} & \textbf{Training Loss} & \textbf{Parameter Count} \\
        \hline 
        SNN : {\it MCLeaky} & 98.8 & 0.062 & $700K$\\
        \hline 
        SNN: {\it QMCLeaky} + QFC & 91.84 & 0.1417 &  $1M$\\ 
        \hline
        SNN: C-LIF & 88.82 & 0.125 & 500K\\
        \hline 
        SNN: {\it DM-SLSTM} & 89.56 & 0.35 & $3M$\\
        \hline 
        SNN: SLSTM\#1 & 86.73 & 0.5667 &  $700K$\\
        \hline
        ANN: LSTM & 95.31 & 0.35 &  $500K$\\
        \hline
        
    \end{tabular}
    } 
    \label{tab:ACC}
\end{table}

From the Table~\ref{tab:ACC}, it was observed that {\it MCLeaky} neuron combined with a fully connected layer performs the best compared to the SOTA LIF or current based LIF (C-LIF) neurons. {\it MCLeaky} managed to deliver a model with a supreme 98.8\% test accuracy, which is enhanced by 8\% and 3\% 
over the SOTA LIF and C-LIF neuron-based SNN models, respectively. 
 This further validates that adding additional memory to the neuron helps while incurring minimal latency impact. Considering the
best candidate model and few other runner-ups, {\it MCLeaky} neuron with more dropouts than LIF and C-LIF stands out from the rest, thereby realizing a better representation of memory structured forget
gate, 
which is similar to an LSTM
architecture. 
Besides {\it MCLeaky} candidates benefit from high compute resource savings compared to the conventional LSTM. 
Figure~\mbox{\ref{fig:EmotionPipeline}}~depicts a low-power, low-latency and an {\em edge} setup solution designed for 
validating the performance of emotion recognition model using
in-house EDA sensor data.



\begin{figure}[hbt!]
    \centering
    \includegraphics[width=0.95\linewidth]{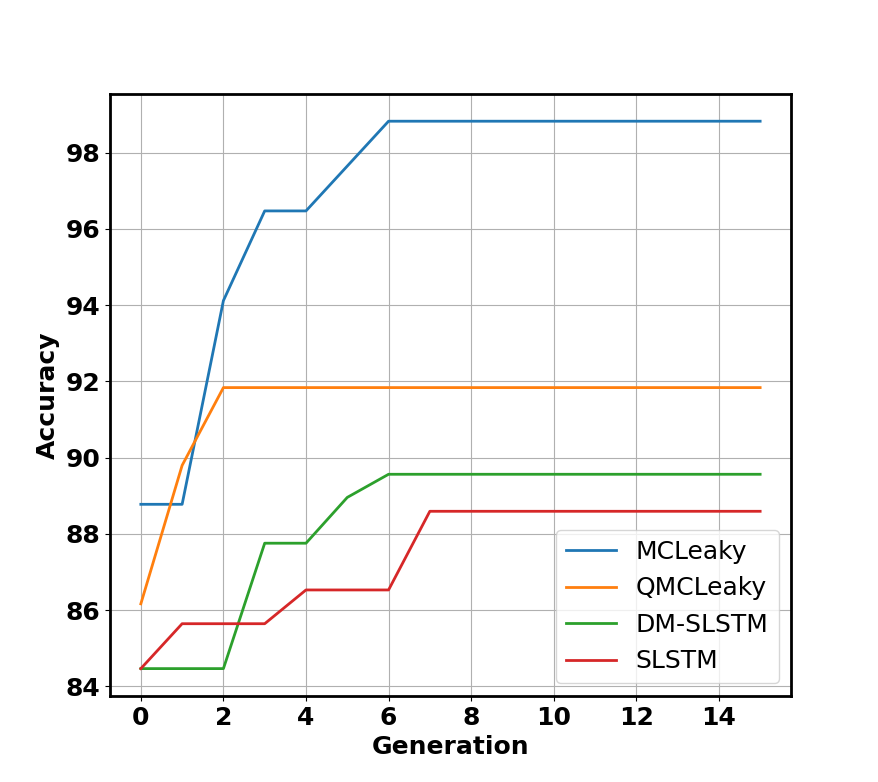}
    \caption{Model Accuracy versus Generations plot of NAS runs for {\it MCLeaky}-SNN, {\it QMCLeaky}-QNN, LIF-{\it DM-SLSTM} and LIF-{\it SLSTM}. }
    \label{fig:accVGen}
\end{figure}

Figure~\ref{fig:accVGen} presents the
maximum accuracy attained by the candidate solutions in each generation during the NAS run.
As expected, the accuracy of the best candidate model extracted from the NAS run, continues to rise in the initial part of the generations and later remains 
constant over generations.
While {\it QMCLeaky} and {\it SLSTMs} converged faster, {\it MCLeaky} model took 13 generations to evolve to its best form, exhibiting the highest test accuracy over others. The {\it QMCLeaky} comes second-best in terms of test accuracy, besides converging in a fewer generations.

Additional tests are conducted using the temperature modality to assess {\it MCLeaky}'s input signal generalizability. This includes training the best network from the prior networks on the new data. These models are built with unimodal temperature wrist data and combinations of multimodal EDA and temperature data. Performance of these models are tabulated in table \mbox{\ref{tab:paperComparison}}.

  

Table~\mbox{\ref{tab:paperComparison}} lists higher performance
of multimodal SNN model with EDA Combo and Temperature data over other single modalities. EDA Combo refers to combination of EDA Chest and EDA wrist data.  The EDA Combo with Temperature performed
better than others while conceding low input neurons.
Less input sample requirements implies that the model has to wait less  for the buffer to fill up, which improves inference latency. Since EDA Combo model requires 128 samples before it starts processing  and EDA Combo + Temperature model needs only 48, the relative inference latency is improved by 2.66~$\times$.
Table~\mbox{\ref{tab:paperComparison}} also presents different modality of signals
including Temperature data, EDA Wrist data, EDA Chest data, ECG data,
and multiple combinations of them. The accuracy of the {\it MCLeaky} neuron incorporated SNN model for four different modality of 
signal towards the same target classes were found acceptable. This validates functioning of the proposed {\it MCLeaky} neuron on 
different signals, and its adaptability for different modalities of time-series signals.

\begin{figure}
    \centering
    \includegraphics[width=\linewidth]{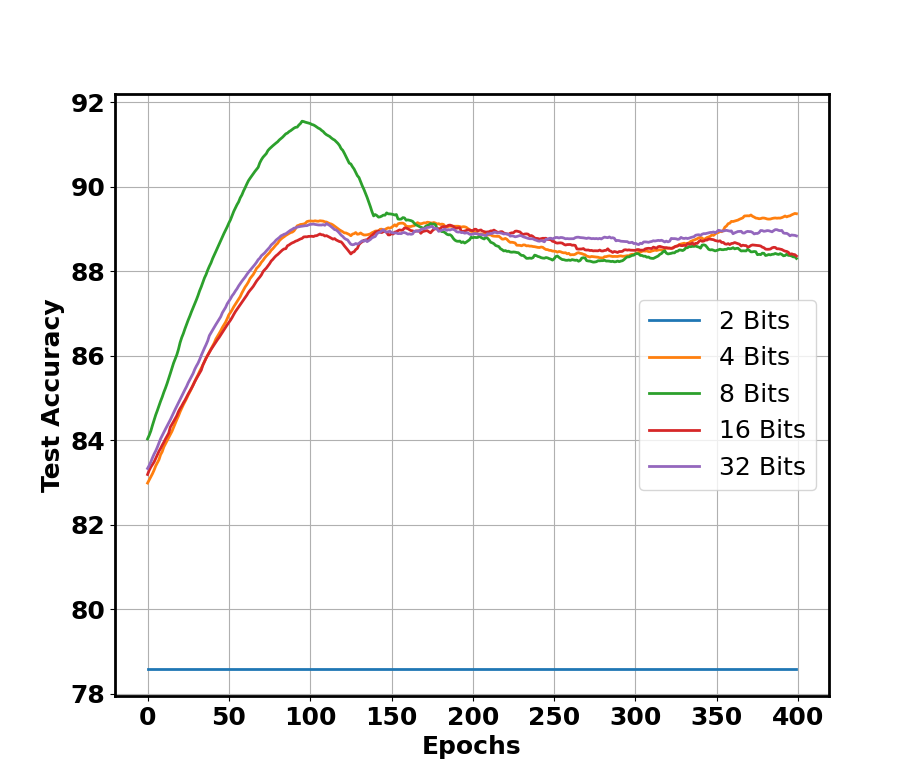}
    \caption{Test accuracy's on training the best {\it QMCLeaky} network with varying quantized bit lengths for 400 epochs. Wavy pattern in the plots occurs due to the use of Cosine Annealing optimization algorithm on learning rate. The optimization scheme was adopted to avoid runs to converge on local minima, 
    a common problem of QNNs due to loss of precision.}
    \label{fig:qBitComparison}
\end{figure}

For evaluating the hardware readiness of {\it MCLeaky} neuron, a {\it QDSNN} was designed using the 
{\it MCLeaky} neuron and fully connected layers.
Quantized layers are incorporated in the NAS framework
to extract the best-quantized model.
Running NAS with {\it QMCLeaky} neuron and quantized fully connected layer resulted in the best model characterized with 91.84\% classification accuracy while consuming over 100~K trainable parameters. An overall 7\% drop in accuracy compared to its non-quantized variant was established. A drop in performance is expected in the field of quantized models with fixed point operations.
Multiple bit lengths, including 2, 4, 8, 16, and 32 were further evaluated on the best-extracted model 
from  NAS runs that are configured for {\it QMCLeaky} neurons. 
The evaluated results for the networks trained for 400 epochs are shown in the Figure~\ref{fig:qBitComparison}. 
 
\mbox{{\it QMCLeaky}} and \mbox{{\it MCLeaky}} networks are also tested with MIT-BIH's ECG dataset to analyse the robustness of the neurons. The best network for MIT-BIH ECG data produced \mbox{\textit{MCLeaky-SNN}} and \mbox{\textit{QMCLeaky-QDSNN}} models with 93.47\% and 92.21\% test accuracy respectively, as listed in Table~\mbox{\ref{tab:paperComparison}}.


Table \ref{tab:hwAccuracy} 
lists the comparison of the proposed work with other hardware implementations of EDA stress detection.
It includes
CNN on the Jetson Nano. Experiments include models that only accept one of the EDA signals either from the chest or from the wrist. This CNN model is compared against other model that is trained for both Chest and Wrist modality of signals.
The CNN model trained on both signals showed an accuracy of around 95\%.
The key points from this comparison include the superior CNN performance for low-frequency EDA Wrist signals on Jetson Nano but a drop in accuracy for higher-frequency EDA Chest signals as presented in Table~\ref{tab:hwAccuracy}.
The quantized version on FPGA presents accuracy metrics in the range of 85.12\% to 91.84\%, a more consistent result range when compared to a drastic drop in performance for higher frequency EDA Chest signal in the case of CNNs.
 The {\it QDSNN} Combo model on PYNQ-Z1 FPGA\footnote{Xilinx Zynq based FPGA board having 650 MHz dual core Cortex-A9 processor, 53K LUTs, 106K flipflops in the integrated programmable logic (PL) fabric with 143~MHz clock and 630 KB fast block RAM, running Theano with Lasagne v0.1.} stands out as the best among the investigated models that run on different hardware platforms.

\begin{table}[!htp]
    \centering
    \caption{Comparison of model performance on various hardware.}   
    
    \begin{tabular}{|c|c|c|}
    \hline
        \textbf{Model} & \textbf{Data} & \textbf{Accuracy} \\
        \hline
        QDSNN FPGA  & EDA Combo & 91.84 \\
        \hline
        QDSNN FPGA  & EDA Wrist & 88.60 \\
        \hline        
        QDSNN FPGA & EDA Chest & 85.12 \\
        \hline                
        CNN Jetson Nano & EDA Combo & 95.31 \\
        \hline
        CNN Jetson Nano & EDA Wrist & 91.30 \\
        \hline
        CNN Jetson Nano & EDA Chest & 82.92 \\
        \hline        

    \end{tabular}
    \label{tab:hwAccuracy}
\end{table}

For CNNs executed on Jetson Nano, {\tt jtop} tool was used to measure power consumed. \footnote{NVIDIA Jetson Nano results use Jetson Nano 4GB board running Jetpack SDK 4.6.1 on Ubuntu 18.04 LTS operating system and CUDA 10.2, power numbers taken from the JTOP tool.}
And for {\it QDSNNs}, synaptic operations are considered for energy comparisons. The Synaptic operations are directly proportional to the number of synapses, representing the number of Multiplier-and-Accumulator (MAC) operations in CNN and the number of accumulator (ACC) operations in SNN. Hence, the energy consumption of CNN and SNN models are expressed as stated in Equations~\ref{EqER}.
Equation~\ref{EqER} mentions the energy consumption ratio of one single MAC in convolutional ANN to an accumulator unit in SNN. 
Therefore the relative energy between SNN and ANN is deduced as stated in the Equation~\ref{EqER}, where $E_{ANN}$ and $E_{SNN}$ are the energy consumption of ANN and SNN respectively, and $E_{MAC}$ and $E_{ACC}$ are the energy cost of one single MAC in ANN and one single accumulator in SNN respectively. 


\begin{equation}
\left.
\begin{split}
E_{ANN} &= N_{MAC} \times E_{MAC} \\
E_{SNN} &= N_{ACC} \times E_{ACC} \\
\lambda &= \frac{E_{MAC}}{E_{ACC}} \\
\frac{E_{SNN}}{E_{ANN}} 
    &= \frac{1}{\lambda} \times \frac{N_{ACC}}{N_{MAC}},
\label{EqER} 
\end{split}
\right\}
\end{equation}

ANN has a single activation per synapse, with input being static and synchronous. Therefore, there is only one MAC operation per synapse. Thus, $N_{MAC}$ is proportional to $N_{synapses}$. However, in SNNs, the number of activations depends on the number of spikes. Synaptic activity in a neuron depends on several hyper-parameters and inputs that are retained for more than one time-step. Hence, $N_{ACC}$ is directly proportional to $N_{spikes}$. In summary, the Equation~\ref{EqER} is further simplified and stated in the Equation~\ref{EqERnew}.
$\lambda$ depends on the target 
technology node employed for tapeout process.
$\lambda \approx 5$ was considered for synthesizing 16-bit adders and multiplier designs in the Cadence Genus tool using 130~nm skywater technology library files. \hlg{N. Rathi, et al.\mbox{~\cite{Rathi2023}}} considered $\lambda$ of $5.11$ for 32-bit design when synthesized on 45~nm gpdk library, specifying that the ratio might vary from one technology to another although it remains greater than one in
most of the cases. Hence, considering higher technology process of 130~nm, we consider 
$\lambda = 5$ for this proposed work. 
\begin{align}
\frac{E_{SNN}}{E_{ANN}} 
    &= \frac{1}{\lambda} \times \frac{N_{spikes}}{N_{synapses}}, 
\label{EqERnew} 
\end{align}
For the fully-connected layers, spike ratio $SR$ is represented as stated in the Equation~\ref{SR_FC}, 
where $N(l-1)_{output\_spikes}$ is the number of output spikes at layer $l-1$, $N(l)_{neurons}$ is the number of neurons at layer $l$. $SR$ is therefore deduced to Equations~\ref{SR_OL}.
\begin{align}
SR(l) &= \frac{ N (l-1)_{output\_spikes} \times N(l)_{neurons} } {N(l-1)_{neurons} \times N(l)_{neurons}}
\label{SR_FC} \\
SR(l) &= \frac{N_{spikes}}{N_{synapses}} = 
\frac{ N (l-1)_{output\_spikes} } 
{N(l-1)_{neurons} },
\label{SR_OL} 
\end{align}

We trained and tested {\it QDSNNs}, SNNs and ANNs models on  the 
aforementioned dataset with Leave One Subject Out (LOSO) for cross-validation of unimodal and multi-modal models. 

\begin{table}[!htp]
    \centering
    \caption{Power and Latency of the ANN model tested on Nvidia Jetson Nano. The latency is for a one data sample, and power represents the steady state power usage as reported in {\it jtop} tool.}
    \resizebox{0.30\textwidth}{!}{
    \label{tab:jetsonPowerLatency}
    \begin{tabular}{|c|c|c|}
    \hline
        Data & Latency~(ms) & Power~(mW) \\
        \hline
        EDA Combo & 2.64 & 204 \\
        \hline
        EDA Wrist & 2.43 & 163 \\
        \hline
        EDA Chest & 3.61 & 365 \\
        \hline
    \end{tabular}
    }

\end{table}

\strutlongstacks{T}
\begin{table*}[!htp]
    \begin{center}
    \caption{Energy Efficiency and EDP gain with $\lambda = 5$ for the best performing NAS evolved candidate architecture blocks and neuron parameters forming the model.}

    \resizebox{0.8\textwidth}{!}{
    \begin{tabular}{|c|c|c|c|c|}
        \hline
            \textbf{Dataset}  &             
            
            \Longstack{ \textbf{MCLeaky Energy Efficiency} 
            \\ 
            \\
            (  $\frac{E_{ANN}}{E_{MCLeaky-SNN}}$ )    
            }  &
            
            \Longstack{ \textbf{LIF-DM-SLSTM Energy Efficiency}
            \\ 
            \\
            (  $\frac{E_{ANN}}{E_{DM-SLSTM}}$ ) 
            }  &
 
            \Longstack{ \textbf{ MCLeaky EDP ratio} 
            \\ 
            \\
            ($\frac{EDP_{ANN}}{EDP_{MCLeaky-SNN}}$ ) }  &
  
            \Longstack{ \textbf{ LIF-DM-LSTM EDP ratio} 
            \\ 
            \\
            ($\frac{EDP_{ANN}}{EDP_{DM-SLSTM}}$ ) }  \\
                    
            \hline
            EDA Combo     & $25.12\times$ & $2.08\times$ & $52.37\times$ & $4.38\times$  \\
            \hline
            
            \hline
            EDA Wrist     & $35\times$ & $3.73\times$ & $74.08\times$ & $7.89\times$  \\             
            \hline
            
            \hline
            EDA Chest     & $39.2\times$ & $5.36\times$ & $81.9\times$ & $11.2\times$  \\             
            \hline

    \end{tabular}}

         \label{table:EnergyResults}
         \end{center}
\end{table*}

Table~\ref{table:EnergyResults} shows Energy Efficiency of the model for the EDA Combo, EDA Wrist and EDA Chest based on the average $SR$ for the entire network. Table~\ref{table:EnergyResults} clearly states the energy gains of 25.12~$\times$, 35~$\times$ and 39.2$\times$ for the respective {\it MCLeaky} enriched SNN models 
over 
the CNN models that are trained on EDA Combo, EDA Wrist and EDA chest datasets respectively.
Whereas, the LIF-based {\it DM-SLSTM} shows energy efficiency of 2.08~$\times$, 3.73~$\times$ and 5.36~$\times$ over the ANN model when trained on EDA Combo, EDA Wrist and EDA Chest datasets respectively. The power and latency values of ANN models tested on Jetson Nano edge-GPU for EDA data from WESAD dataset are listed in Table~\ref{tab:jetsonPowerLatency}.
On the MIT-BIH ECG dataset, \mbox{{\it MCLeaky-SNN}} rendered superior accuracy of $93.47\%$ and has an energy efficiency of $24.9~\times$ over ANN. The bio-realistic adaptive exponential (AdExp) LIF neuron model on DYNAP-SE ~\cite{Azghadi2020}, a mixed-signal hardware has an energy consumption of $17pJ$. Though the Adaptive Exponential LIF neurons are biologically realistic and low power, they are computationally  expensive to realize it on hardware,\hlg{ as presented in Xiao, et al.\mbox{~\cite{Xiao2020}}}.

\begin{figure}[htp!] 
\centering
\includegraphics[scale=0.30]{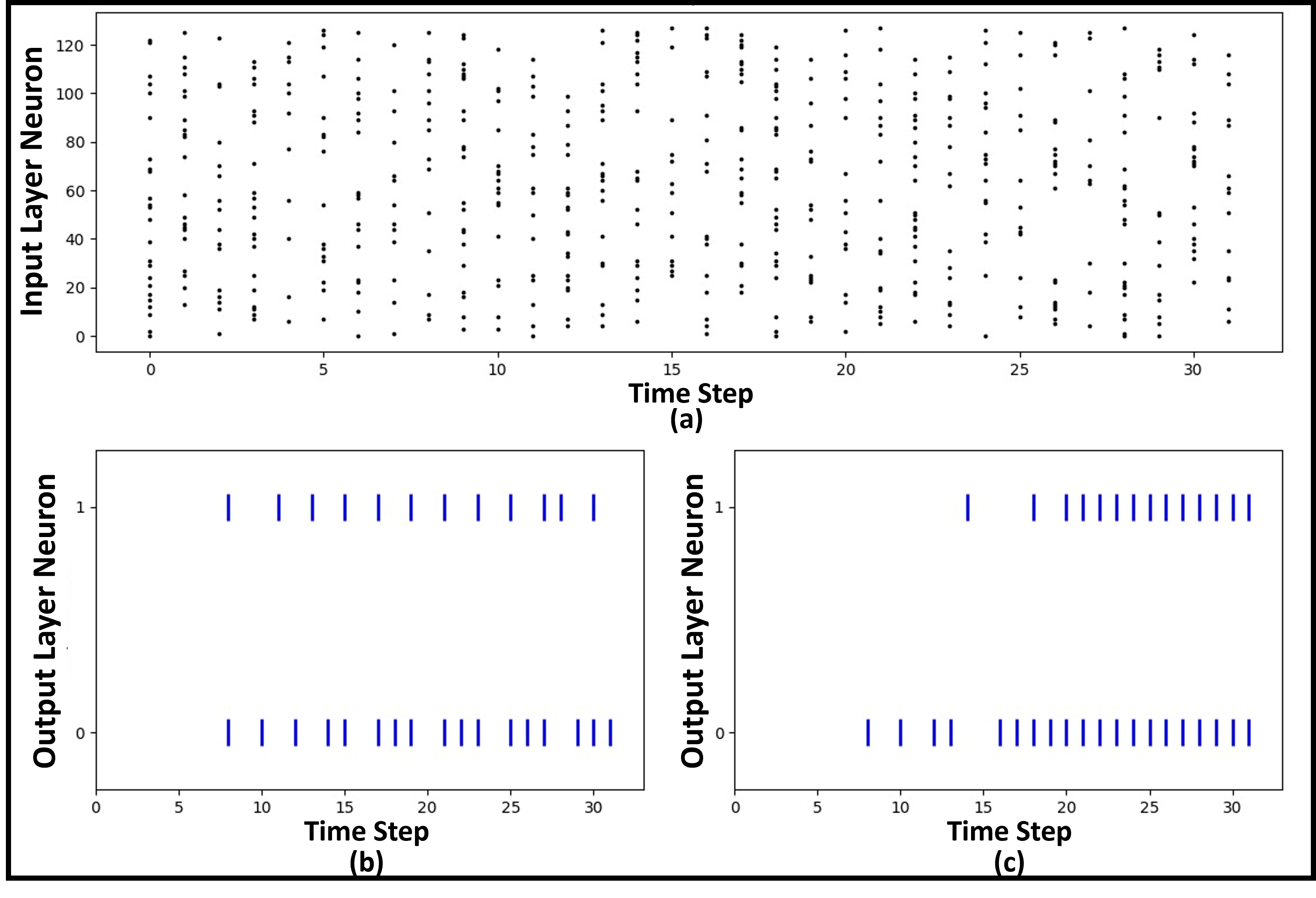}
\captionsetup{justification=centering}
\caption{Raster plot for (a) Input Layer neurons, (b)  {\it MCLeaky-SNN} Output Layer neurons and (c) {\it LIF-DM-LSTM}  Output Layer neurons. Every dot and line  represents one spike emitted by a neuron in the input and output layer respectively.}
\label{fig:RasterPlot}
\end{figure}

The somatic firing for the Output layer which 
includes 2 output neurons
in {\it MCLeaky-SNN}, and {\it LIF-DM-LSTM} networks
that is applied for EDA Wrist test data is presented 
in Figure~\ref{fig:RasterPlot}.
It reveals that the {\it MCLeaky-SNN} model processes the neural information using very sparse neuronal spikes, which results in low power consumption.

\section{Conclusion and Future Work}

In summary, 
the {\it MCLeaky} neuron introduced in this work outperforms the best form of spiking LSTM model generated from the NAS runs by 8\%.
Additional comparison among the {\it SLSTMs} 
suggests that although 
approximating few components of the gate functions offers parameter savings,
in the medical domain where the detection is critical, 
{\it SLSTM} and {\it DM-SLSTM} offer high performance, which is further bettered by {\it MCLeaky} SNN network. {\it MCLeaky} neurons are evaluated for diverse datasets and are found to be energy efficient, achieving superior accuracy.  
The accuracy of the {\it MCLeaky} neuron incorporated SNN model for four different modality of 
signals including EDA Chest signal, EDA Wrist signal, Temperature data, ECG signal 
towards the same target classes were found acceptable. This proved
that {\it MCLeaky} neuron is highly adaptable for different modalities of time-series signals.
A comparison with prior SOTA that 
works on signal pre-processing, and feature extraction approach,
is considered an additional effort, while
the proposed work effectively operates on the normalized raw signals.
{\it Quantized MCLeaky} neuron incorporated in {\it QDSNN} presents low-power 
and low-latency solutions for stress detection, which is a step towards building accurate and real-time emotion recognition 
system on the edge computing device using 
physiological signals.
The paper presents hardware comparisons of 
ANN model on Jetson Nano and SNN on PYNQ FPGA board.

The proposed {\it MC-QDSNN} on FPGA shows comparable performance metric with the best ANN model on Jetson Nano GPU. However the {\it QMCLeaky-QDSNN} on FPGA is preferred over other implementations owing to large power savings and minimal inference latency characteristics. 
In summary, {\it MCLeaky-SNN} offers energy savings in the range of 25.12$\times$ to 39.2$\times$, and EDP gain
in the range of 52.37$\times$ to 81.9$\times$ for different modalities of EDA signal over ANN model. Additionally, 
{\it MCLeaky} based SNN offers the best accuracy of 98.8\% which 
stands-out among other implementations 
including 
ANN and its variants, and SLSTM and its corresponding variants.
The introduction of {\it MCLeaky} neuron to realize the memory component for the SNN models is thus found effective and is a best candidate to realize long short term memory for Spiking inputs.

\bibliographystyle{unsrt}
\bibliography{arxiv.bib}


\end{document}